\documentclass{article} 
\usepackage{arxiv}
\usepackage{amsmath}
\usepackage{times}

\usepackage[hidelinks]{hyperref}
\usepackage{xurl}
\usepackage[normalem]{ulem}

\usepackage{array}
\newcolumntype{L}[1]{>{\raggedright\let\newline\\\arraybackslash\hspace{0pt}}m{#1}}
\newcolumntype{C}[1]{>{\centering\let\newline\\\arraybackslash\hspace{0pt}}m{#1}}
\newcolumntype{R}[1]{>{\raggedleft\let\newline\\\arraybackslash\hspace{0pt}}m{#1}}

\usepackage{xspace}
\usepackage{amsthm}
\usepackage{graphicx}
\usepackage{booktabs} 
\usepackage{adjustbox}
\usepackage{siunitx}
\usepackage{bm}
\usepackage[hashEnumerators,smartEllipses,hybrid]{markdown}

\newcommand{\ie}{i.e.~}
\newcommand{\eg}{e.g.~}
\newcommand{\egnows}{e.g.} 
\newcommand{\wrt}{w.r.t.\ }

\newcommand{\secsummary}{\subsection{Results and remarks}}

\iclrfinalcopy 

\title{Tailored Uncertainty Estimation for\\ Deep Learning Systems}

\author{
    \small{Joachim Sicking$^{1,2}$, Maram Akila$^{1}$, Jan David Schneider$^{3}$, Fabian Hüger$^{3}$,}\vspace{0.05cm}\\ \small{\hspace{0.1cm}\textbf{Peter Schlicht}$^{3}$, \textbf{Tim Wirtz}$^{1,2}$, \textbf{Stefan Wrobel}$^{1,4}$}\vspace{0.1cm}\\
    $^1$ \small{Fraunhofer IAIS}, $^2$ \small{Fraunhofer Center for Machine Learning}, \\
    $^3$ \small{Volkswagen AG}, $^4$ \small{University of Bonn}\vspace{0.1cm}\\
    \scriptsize{\texttt{\hspace{0.1cm}\{joachim.sicking, maram.akila, tim.wirtz\}@iais.fraunhofer.de,}} \\
    \scriptsize{\texttt{\hspace{0.1cm}\{jan.david.schneider, fabian.hueger, peter.schlicht\}@volkswagen.de,}}\\
    \scriptsize{\texttt{\hspace{0.26cm}stefan.wrobel@cs.uni-bonn.de}}\\
}

\begin{document}

\maketitle


\begin{abstract}
Uncertainty estimation bears the potential to make deep learning (DL) systems more reliable. 
Standard techniques for uncertainty estimation, however, come along with specific combinations of strengths and weaknesses, \eg with respect to estimation quality, generalization abilities and computational complexity. 
To actually harness the potential of uncertainty quantification, estimators are required whose properties closely match the requirements of a given use case.
In this work, we propose a framework that, firstly, structures and shapes these requirements, secondly, guides the selection of a suitable uncertainty estimation method and, thirdly, provides strategies to validate this choice and to uncover structural weaknesses.
By contributing \textit{tailored} uncertainty estimation in this sense, our framework helps to foster trustworthy DL systems.
Moreover, it anticipates prospective machine learning regulations that require, \eg in the EU, evidences for the technical appropriateness of machine learning systems.
Our framework provides such evidences for system components modeling uncertainty.
\end{abstract}


\section{Introduction}
\label{sec:intro}

Uncertainty estimation in neural networks aims at telling apart two types of statistical concepts: 
those that the model learned during its optimization and those it did not ``internalize''. 
A proper uncertainty estimator ``recognizes'' if the statistical concepts of unseen data inputs are (un-)familiar to the network and thus how trustworthy the respective network outputs are. 
Uncertainty quantification is in that sense about constructing accurate ``self-assessment'' mechanisms for models that are learned from data. 
This renders them a cornerstone of \textit{trustworthy (safe) machine learning}, a sub-field of machine learning (ML) that tackles model insufficiencies, such as lack of reliability, which prevent recent neural networks, despite their potential, from being deployed in safety-critical applications (see \eg \citet{amodei2016concrete,koopman2018heavy,houben2021inspect}). 
While larger training sets may help to reduce the model's uncertainty \citep{vapnik2000statistical} and thus the need to accurately model it, we can still not hope for data to ``come to the rescue'' given the sheer size of the model input spaces 
(especially for real-world applications)
and irreducible data noise. 

Let us exemplify the practical relevance of uncertainty estimation by introducing two central classes of applications: 
\textit{uncertainty for safety} (\eg \citet{le2018odsafety,shafaei2018safety,aravantinos2020uncertainty}) and \textit{uncertainty for (semi-)automation} (\eg \citet{thrun2005robotics,giannetti2017uncertaintyindustry,sunderhauf2018dlforrobotics,papananias2019manufacturing}), respectively. 
The already above mentioned safety-critical applications (uncertainty for safety) comprise use cases that range from recommendation engines, \eg in the medical context, to cyber-physical systems such as robots, drones or vehicles. 
A domain expert's notion of a ``good'' uncertainty mechanism in these fields is likely given by a high-level requirement, \eg ``do not provide a semantic segmentation of a computed tomography (CT) image unless you are 99.5\% certain about it" (in a medical application) or even more abstract such as ``contribute to defensive driving and compliance with the traffic law" (in the case of an automated vehicle).

Apart from these safety-critical applications, uncertainty quantification 
can be important
for \mbox{(semi-)auto}\-mation, \eg when phasing in ML systems for quality assurance (QA) in an industrial production plant or to (partially) automate the classification of texts \citep{lewis1994text}. 
For QA purposes, an image-processing model could decide whether or not the surface of a (mass) product is sufficiently homogeneous and undamaged. 
If the model is uncertain regarding its decision, the product is inspected by a human expert to ensure high quality standards.
A practitioners' requirement regarding this QA model might be that ``uncertainty estimates allow to flexibly maximize the degree of automated quality inspection such that less than $1$ in $10,\!000$ products leaves the plant despite being in a defect state". 
In case of the automated text annotator one might similarly ask for ``accurate uncertainty estimates that allow to hand over difficult examples to a human annotator such that the results of the ML annotation tool (with occasional human overrule) have the same quality as the ones of a fully manual annotation process". 

This 
practitioner's view on uncertainty estimators varies from the perspective of ML researchers putting forward novel concepts to modeling and testing uncertainty estimates. 
Not only does a complexity gap occur between lab-oriented scientific proposals and actual industrial applications,
the practitioner's focus is moreover on the alignment of the actual technical state-of-the-art with a ``desired state'', \ie the respective application requirements.
Such a systematic alignment is a prerequisite for responsibly deploying ML systems, \ie in a way that mitigates their risks and thus allows their strengths to play out safely.
We contribute to this alignment by providing a \textit{framework for 
the 
development
and testing of neural uncertainty estimators based on high-level requirements}, aiming at ML applications with tailored uncertainty mechanisms. 
By doing this, we bridge the gap
between, on the one hand, industry researchers and safety engineers who develop standards and set high-level requirements for safe ML such as the sub-committee 42 of ISO's and IEC's joint technical committee 1 \citep{iso2017ai}
and, on the other hand, scientific ML researchers developing uncertainty mechanisms and testing algorithms.
To the best of our knowledge, this is the first work that systematically links these two perspectives on neural uncertainty estimation. 

Developing and testing learned neural models is qualitatively different from developing and testing (traditional) complex systems that are manually assembled. This becomes evident when comparing three complex objects, each composed of tens of thousands smaller units: 
a car, programmed software and a neural network. While we understand the role of a battery, an electric motor and tires in a car (and the classes and objects in an object-oriented software solution) and how they interplay, such knowledge about the ``semantic'' inner workings is inherently inaccessible for a neural network. 
It is a non-interpretable black box that is optimized using meta-algorithms. 

These properties of neural models necessitate novel approaches to testing. While traditional hardware and software systems have been specified \citep{dwyer1999finitestateverification,tahat2001reqbasedtestgeneration,grunske2008specificationpatterns,pandey2010requirementeng,meth2015designingrequirement} and tested \citep{kropp1998robusttesting,tahat2001reqbasedtestgeneration,nebut2003productfamilies,majzoobi2008hardwaresecurity,klein2009verifyoskernel,kavitha2010requirementbased,esteve2012satellite} for decades, sometimes even with the help of mathematical proofs \citep{hartman2006testing},
respective approaches (often) rely on splitting a complex system into smaller ``semantic'' units (\eg unit testing to debug software).
Smaller units are easier to safeguard, their roles (in a logical or physics-based cause-and-effect model) and failures are easier to understand and can either be mitigated by understanding (and removing) the cause or by adding redundancy as a fail-safe. 
A deep neural network (DNN) on the contrary is not composed of semantic building blocks and thus does not become ``safer'' by adding an additional (or redundant) layer.
Moreover, the (effective) input spaces in classical engineering (\eg the forces and torques a machine part is exposed to) are small compared to, \egnows, the (often) million-dimensional input spaces an image-processing ML model operates on. 
Such smaller input spaces allow for more comprehensive testing, \ie sampling the input space in ways such that most application scenarios can be seen as interpolations of these sample points. 
Furthermore, the physics-based (or logical) models that describe such assembled systems ``ensure'' a certain stability and thus predictability of the system behavior in these interpolation cases. 
This is in contrast to (many) ML systems that are optimized on extremely sparse data samples of high-dimensional spaces such that distinctions between inter- and extrapolation are hardly possible. 
These purely data-based models 
provide moreover less stable generalization as their outputs are highly susceptible to certain types of small changes in input space (see adversarial attacks \citep{goodfellow2014adversarialexamples}). 

This combination of non-interpretable models and high-dimensional input spaces
renders it hardly possible to exhaustively test semantic and high-level requirements for real-world ML models.
Nevertheless, \textit{testing does yield insights into multiple aspects of the qualitative and quantitative behavior of an ML function} within a given computing and cost budget. 

While these considerations hold for neural models in general, they also apply in the special case of developing and testing neural uncertainty estimators that are integrated into deep learning (DL) models. 
Accurate uncertainty estimators provide an implicit description of the model's domain of proper functioning which makes them promising tools for safe ML. 
Testing uncertainties therefore differs from ML performance testing (\eg of a specifically designed layout of a hidden layer) not only by the considered metrics (\eg uncertainty calibration measures that are ``orthogonal'' to performance metrics) but also in the data selection strategies (by putting a focus on transitions to out-of-data or safety-critical scenarios).

In this work, we provide a framework for systematically breaking down high-level requirements onto uncertainty modeling techniques and uncertainty test cases.
Following the top-down structure of the framework ensures compatibility with many requirement-based development approaches used also for non-ML systems and components.
In detail, we contribute
\begin{itemize}
    \item by categorizing potential requirements, for instance \wrt uncertainty quality, generalization ability (\eg extrapolation) or technical aspects (\eg when building upon \mbox{previous systems}),
    \item by analyzing how and to which extend requirements can be addressed by the differing paradigms of uncertainty modeling and thus, which type of uncertainty estimator to choose for a given DL use case,
    \item by providing a hierarchy of model tests to evaluate whether the uncertainty requirements are met by the selected estimator, and whether systematic weaknesses exist which might hinder its further use or deployment.
\end{itemize}

The paper is organized as follows: first, we contextualize our work in the fields of trustworthy ML and ML testing in section~\ref{sec:related}. Next, we lay out the structure of our uncertainty
framework
in section~\ref{sec:process}. Caveats regarding neural uncertainty estimators and best practices on how to address them are presented for each of the five steps of the framework in sections~\ref{sec:initial_demand} to \ref{sec:tests}. An outlook in section~\ref{sec:discussion} concludes the work.


\section{Related work}
\label{sec:related}

As a technical tool to establish safe ML systems, uncertainty estimation may form a cornerstone to satisfy
demands of upcoming ML regulations and standards some of which are outlined in the first ``paragraph''.
Next, we sketch already existing approaches to operationalize trustworthy ML as well as challenges that practitioners face when building reliable ML systems (second ``paragraph'').
Subsequently, algorithmic techniques that increase the reliability of learned systems are reviewed (third ``paragraph'').
An important class of algorithmic approaches are neural uncertainty estimators. We present their technical foundations, namely modeling techniques and evaluation schemes, in the last ``paragraph''.


\paragraph{Trustworthy ML from a regulatory perspective}
Both official institutions and industry are about to set rules for ML systems. 
Aiming for an effective protection of basic rights in an era of ubiquitous ML systems, several countries put forward law initiatives and guidelines for regulating ML models. 
In November 2020, the White House released guidelines for public authorities in the US on how to regulate artificial intelligence (AI) systems \citep{whitehouse2020guidance}. 
In April 2021, the EU commission published a ``proposal for a regulation laying down harmonized rules on AI" \citep{eu2021regulation}) that (among others) classifies and handles ML applications according to their ``criticality''. 
The proposal moreover asks for technical documentations that detail how such critical ML systems function.
Aside these official regulations (and, if applicable, in accordance with them), industry and standardization bodies define minimal requirements, interfaces and liability regimes as prerequisites for functioning technological markets. 
International standards are \eg developed by the sub-committee 42 of ISO's and IEC's joint technical committee 1 \citep{iso2017ai} that issued technical reports \eg on the trustworthiness in AI (ISO/IEC TR 24028:2020, \citet{iso2020trustworthyai}) and on the assessment of the robustness of neural networks (ISO/IEC TR 24029-1:2021, \citet{iso2021assessrobustness}). Actual standards \eg for AI management systems (ISO/IEC CD 42001, \citet{iso2021managementai})
are under way.
These international initiatives build on standardization efforts on national level, see \eg the German standardization roadmap on AI \citep{din2020roadmap}, and harmonize them.
Moreover, application-specific norms come into existence, for instance, the ANSI/UL 4600 standard for safety for the evaluation of autonomous products \citep{ul2020autonomous} that, among others, targets applications in the field of autonomous mobility.
For the specific context of safe AI for road vehicles, see \citet{iso2021safe-ai-road-vehicle}.


\paragraph{Trustworthy ML from a practitioners' perspective}

When developing industry-scale ML applications from (newly invented) ML modeling techniques, corporate researchers and practitioners face challenges that are often not addressed by basic research in machine learning: \cite{holstein2019improving}, for instance, survey ML product teams \wrt to fairness in ML and identify challenges such as the unavailability of both high-quality datasets and tools for fairness-focused debugging.
Broadening the focus,
\cite{lwakatare2020large} provide a literature review of obstacles such as adaptability and scalability that ML practitioners encounter when developing and maintaining ML-based systems. 
Similarly, \cite{ishikawa2019engineers} survey ML engineers to identify ML-specific challenges from a software-engineering perspective.
\cite{sculley2015hidden} compile ML-specific risk factors, \eg boundary erosion and hidden feedback loops, that are likely to cause technical debt when operating application-scale ML models. 

At the same time, applied researchers and practitioners propose hands-on approaches to operationalize trustworthy ML. For instance, the concept of model reporting by means of model cards \citep{mitchell2019cards}, and comparably, the idea of providing fact sheets that outline relevant attributes of an ML service \citep{arnold2019factsheets}.
Operationalizations of trustworthy ML for safety-critical systems take these approaches a step further and require specific (statistical) evidences (\eg quantitative tests) that are bound together by an overarching safety argumentation which motivates their setup and configuration \citep{koopman2019standard,mock2021safetyargumentation}. 
An important example for such structured, qualitative argumentations are so-called safety assurance cases (that have their roots in classical safety engineering, see ISO standard 15026 \citep{iso2019assurance}). 
For concrete approaches to mitigate safety concerns (and thus contributing to safety argumentations), see \mbox{\eg \citet{koopman2019safetyac,willers2020safety}}.


\paragraph{Algorithmic approaches to trustworthy ML} 

On a technical level, the trustworthiness and reliability of models is fostered by a broad range of machine learning techniques.
Apart from uncertainty estimators, that are discussed below, this entails: interpretability methods (on the level of pixels \citep{selvaraju2017grad,brendel2019approximating} or semantic concepts \citep{kim2018interpretability}), mechanisms to enhance (individual or group) fairness (\eg \citet{adel2019fairness}) and techniques to reduce the susceptibility of ML models \wrt adversarial attacks (\eg by means of robust training \citep{madry2017towards}). 
For a survey on practical methods for ML safety, see \eg \citet{houben2021inspect}.

Systematically evaluating the effectiveness of such safe-ML components is crucial to determine whether they are sufficient \wrt high-level requirements. 
Extensive surveys of the broad body of work on ML testing are provided by \cite{zhang2020machine}, \cite{riccio2020testing} and \cite{braiek2020testing}, respectively.
\cite{zhang2020machine} analyze testing properties, testing components and testing workflows and identify (among others) a lack of research on how different assessment metrics correlate with one another and with the test's ability to uncover faults of the model, respectively.
Numerous algorithmic tools to unveil (and debug) model weaknesses were put forward, \eg DeepXplore \citep{pei2017deepxplore} and TensorFuzz \citep{odena2019tensorfuzz} both of which adapt coverage-guided techniques from software testing to neural models. 


\paragraph{Uncertainty estimation in neural networks}
It is common to distinguish uncertainties by their source. 
The two main sources are data-intrinsic (or aleatoric) uncertainty (\eg a dice roll or a noisy measurement) and model weight (or epistemic) uncertainty which is due to a lack of training data. 
They are complemented by model class risk.  
For a detailed taxonomy of uncertainty types, see \eg \cite{huellermeier2021taxonomy}.  

Several paradigms of modeling such uncertainties exist, each of which containing a multitude of concrete uncertainty modeling techniques (see \eg \cite{kabir2018neural,gawlikowski2021survey,abdar2021review} for extensive reviews). 
One group of approaches directly models uncertainty estimates as network outputs using softmax statistics (for classification tasks, see \eg \cite{guo2017calibration}) and confidence scores (for regression tasks, see \eg \cite{wu2017squeezedet}) or by means of distributional parameters such as Gaussian parametric uncertainty \citep{nix1994estimating}, evidential regression \citep{amini2020evidential} or prior networks \citep{malinin2018predictive}. 
Most of these direct modeling approaches target data-intrinsic (aleatoric) uncertainty. 
Implicit (or non-parametric) uncertainty estimation in contrast relies either on local modifications of model parameters (\eg Monte-Carlo (MC) dropout \citep{gal2016dropout} or SWAG \citep{maddox2019simple}) or on ensembling (\eg deep ensembles \citep{lakshminarayanan2017simple}, mixtures like masksembles \citep{durasov2021masksembles} exist. These approaches typically approximate epistemic uncertainty.
State-of-the-art approaches often employ combinations of parametric and non-parametric approaches to flexibly handle compounds of aleatoric and epistemic uncertainty.

Since explicit uncertainty labels are rarely available, point-wise uncertainty quality can often only be determined relative to the model prediction using \eg the negative log-likelihood (NLL) \citep{nix1994estimating}. 
It is a hybrid between performance and uncertainty quality measure, fulfilling the properties of a proper scoring rule \citep{gneiting2007strictly}. 
Most ``pure'' uncertainty measures however are statistics over (sub-)datasets and quantify how well the distribution of uncertainty estimates matches the distribution of actually occurring model errors. 
To this end, the fractions of absolute model errors and uncertainty estimates are considered, the so-called \textit{normalized residuals}. Building on them, the expected calibration error (ECE, see \eg \cite{guo2017calibration} and \cite{Kuleshov2018}) and its variants such as the marginal calibration error (MCE, \cite{kumar2019verified}) for (multi-class) classification are widely applied to determine calibration quality. 
Other measures focus on specific aspects of the uncertainty estimate distribution, \eg the expected tail loss (ETL, \cite{rockafellar2002conditional}) that allows to analyze worst cases of uncertainty quality \citep{sicking2021wdrop}, a relevant property for many safety-critical use cases.
Further properties of the outlined uncertainty modeling techniques and evaluation measures are discussed in sections~\ref{sec:acceptance_criteria} \mbox{and \ref{sec:mechanism}, respectively.}

A stepping stone towards reliable uncertainty quantification are publicly available open-source code repositories on neural uncertainties like \textit{UQ360} \citep{ghosh2021uq360} and \textit{uncertainty-toolbox} \citep{chung2021uncertainty}. They ease the implementation and evaluation of uncertainty estimators and foster their fair comparison, see \citet{pintz2022survey-unc-toolkits} for a review. While being promising, for now, these frameworks do not contain methods that are specifically developed for more complex architectures like convolutional and recurrent DNNs and support
task-specific approaches and measures (\eg for autonomous driving (AD)) only in a \mbox{limited fashion.}


\section{A framework for developing and testing neural uncertainty estimators}
\label{sec:process}

\begin{figure}[t]
    \centering
    \includegraphics[width=1.00\textwidth]{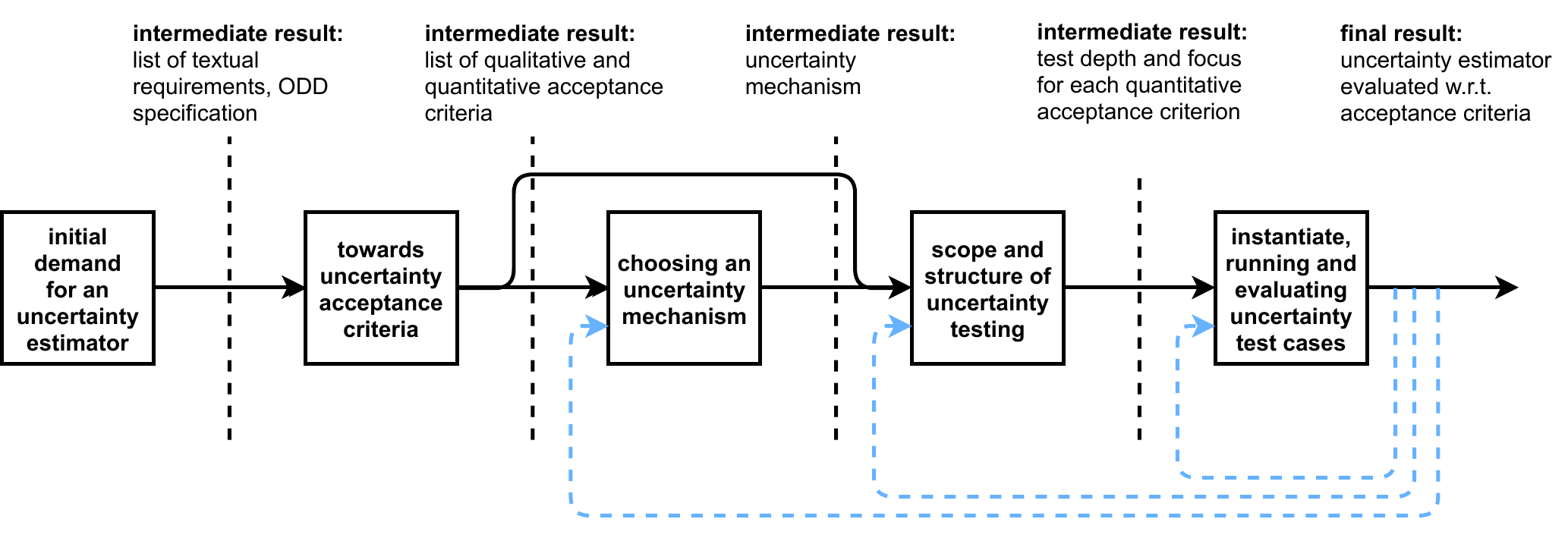}
    \caption{A structured approach to obtain an uncertainty estimator that is tailored to a given deep learning application. The proposed framework is subdivided into five steps in which uncertainty requirements are collected and quantified (steps~1 and 2), an uncertainty mechanism is selected or constructed (step~3) and the quality of its uncertainty estimates is systematically tested (steps~4 \mbox{and 5}), see the following sections for details. The uncertainty acceptance criteria (step~2) provide the basis for both uncertainty modeling (step~3) and test strategy (step~4). Inconsistent or failing tests may entail changes of the test cases, of the testing strategy or of the uncertainty mechanism (see light blue dashed backward arrows).}
    \label{fig:process_diagram}
\end{figure}

While established frameworks for requirement-based development and testing (see \eg \citet{iso2014softwarequality,iso2018functionalsafety}) provide a hull
that remains applicable to ML systems, these learned models require novel technical realizations (see introduction).
For the task of estimating uncertainties in neural networks, we analyze these conceptual challenges and propose best practices for tackling them. 
While the choice of the DNN architecture is, of course, central to the successful utilization of ML, we focus on the ``self-assessment'' capabilities of the learning system, as they are, to a certain degree, more agnostic to the task at hand.
Based on our practical observations, the features and development processes of ML models are typically driven from two sides: 
on the one hand, from an application-oriented direction governed by the needs and requirements of a given (or future) product. 
On the other hand, by dedicated ML researchers and developers who are focused on feasibility and the state-of-the-art in the field. 
The proposed development and testing framework is intended as a structural aid for the interplay between these points of view.

For our purposes, subdividing the framework into five steps provides sufficient granularity. 
These steps and their intermediate results are visualized as a flow chart in Fig.~\ref{fig:process_diagram}. 
In the following, we describe the purpose of each step and name some of their key elements:

\begin{enumerate}
    \item First, we grasp the \textit{initial demand for an uncertainty estimator} by means of high-level (\eg textual) requirements. Understanding the intended purpose as well as the technical modalities of the underlying
    ML system poses the basis for uncertainty modeling and testing. Moreover, the operational domain of the uncertainty estimator is specified, \ie the set of inputs on which it is supposed to function properly.
    \item Next, we structure and refine the high-level requirements by means of requirement categories that we lay out.
    Heading towards \textit{measurable (and thus testable) acceptance criteria}, the quantitative requirements are ``translated'' into 3-tuples of semantic data specification, measure of uncertainty quality and threshold value.
    \item We then \textit{choose an uncertainty mechanism} that seems suitable to meet the acceptance criteria, based on mappings of mechanism-specific characteristics onto the requirement categories.
    Apart from the complexity of the mechanism (``off-the-shelf'' or custom), its ability to model the predominant types of uncertainty is a key property.
    \item Given both uncertainty acceptance criteria and uncertainty mechanism, we \textit{set the scope and structure of uncertainty testing}. 
    To this end, we introduce a hierarchy of tests that builds on semantic and non-semantic data selection strategies. For each acceptance criterion, a testing ``depth'' in this hierarchy is determined and focus points of testing are chosen.
    \item Based on the selected testing focuses for each acceptance criterion, we finally \textit{instantiate, run and evaluate uncertainty test cases}.
    This requires to select concrete test datasets and \eg specific search strategies and their initial parameters.
    Once executed, the binary results of the test cases are aggregated to decide whether the uncertainty estimator fulfills the uncertainty acceptance criteria. 
\end{enumerate}

We detail on each of the five steps of the framework in the following sections.


\section{Initial demand for an uncertainty estimator}
\label{sec:initial_demand}

Many factors, for instance regulatory contexts and technical specifications, influence uncertainty modeling. 
In subsection \ref{subsec:purpose}, we study how these boundary conditions affect 
(directly or indirectly) 
the desired properties of an uncertainty estimator.
The underlying ML use case moreover determines the operational design domain (ODD), i.e. the sets of inputs on which the uncertainty estimator is supposed to function
(see subsection \ref{subsec:modal_and_domain}). 
These input sets are described by semantic dimensions that need to be identified.
Technical specifications may pose additional constraints, \eg on the uncertainty estimator's depth of integration into the ML model (see subsection \ref{subsec:modeling_unc}).
These constraints and the desired uncertainty properties from subsection \ref{subsec:purpose} pose requirements that guide the choice of the uncertainty estimation technique in a later step of the framework.


\subsection{Purpose and desired properties of an uncertainty estimator} 
\label{subsec:purpose}

We broadly distinguish between external and internal purposes of an uncertainty estimator. External purposes range from regulatory requirements over industry standards to company guidelines. Alternatively, they may originate from a surrounding ML system or the need for backward compatibility. Such external purposes are likely to pose additional constraints on modeling and validating uncertainty functionalities. One may think of regulations of the financial industry where detailed technical guidelines for quantitative modeling exist, limiting \eg the use of complex model classes. 

Internal purposes in contrast may be the goal to reach better functional performance (uncertainty for performance) or to employ well-calibrated uncertainty ``sensors'' that could be used to propagate uncertainty information along a chain of ML models or to initiate a hand-over to a backup system or a human (remote) operator. Such internal purposes likely go along with less restrictions on modeling compared to external purposes. 

Uncertainty estimators are moreover employed as \textit{passive} or \textit{active} system components. Passive components monitor the state of the ML system, collecting information that is \eg used for long-term system optimization without having a direct impact in the recent situation. Active uncertainty components, on the contrary, may trigger subsequent actions or changes of the system behavior. The desired properties of these types of uncertainty estimators may vary accordingly: for the passive observer, a high sensitivity to abrupt changes or extreme events might be less relevant as long as the uncertainty estimates are \textit{on average} accurate and bias-free. A dynamic bias, however, might be welcome for an active uncertainty component as \eg a slight overestimation of the uncertainty could provide an additional safety margin. While such an uncertainty actor might be inactive for most input scenarios, it is expected to act accurately in \textit{individual} critical scenarios (as opposed to an on-average accurate behavior, see above). 
In case uncertainty estimates are used to safeguard against several types of undesired model behavior (\eg false positives and false negatives in a classification setting), the individual ``risk profile'' of the use case determines how to balance them appropriately.


\subsection{Operational design domain of the uncertainty estimator}
\label{subsec:modal_and_domain}

Next, we approach the intended \textit{operational design domain} (ODD) of the uncertainty mechanism, \ie the areas of input space where it is supposed to work reliably (see \eg \citet{nhtsa2017adsafety,koopman2019odd}). 
Specifying these areas is especially challenging in open-world settings where input data is not fully under the control of the operator.
In the case of AD systems, for instance, examples for under-specified ODDs could be geographical ones like fixed routes (\eg of a regular bus), lanes exclusively for automated traffic or geo-fenced areas as well as ``situational'' ones like driving under stop-and-go traffic conditions.
However, specifying an ODD in a complex high-dimensional space is not only approximate and incomplete by its very nature, the specifications may moreover have varying degrees of detail and are sometimes even contradictory.

To at least partially account for this fundamental problem, we propose to determine the ODD of an uncertainty estimator by using two techniques that complement each other and that furthermore may allow to identify (potentially occurring) contradictions and inconsistencies: 
on the one hand, by setting semantic boundaries and, on the other hand, by a scenario-based (or edge-case-based) approach where various in-domain-, out-of-domain- and ``borderline'' input scenarios are compiled.
The former approach seeks to describe and threshold relevant semantic dimensions of the input space. 
Taking, for instance, an ML application that processes traffic images, such descriptions comprise scene parameters ranging from the type of traffic over lighting conditions to numbers and positions of traffic participants.
One may furthermore build on existing ontologies or knowledge graphs in the considered field of application if such representations are available and appropriate.
The latter technique builds on collections of in-domain and out-of-domain input scenarios as an exploratory tool to sharpen and improve on the ODD as constructed above. 
Comparing these scenario sets with the identified semantic boundaries, may reveal wrongly or unspecified semantic dimensions, \eg by analyzing whether the selected ``borderline'' scenarios are consistent with the chosen semantic thresholds.

\begin{figure}[t]
    \centering
    \includegraphics[trim=130 100 130 180, clip, width=0.95\textwidth]{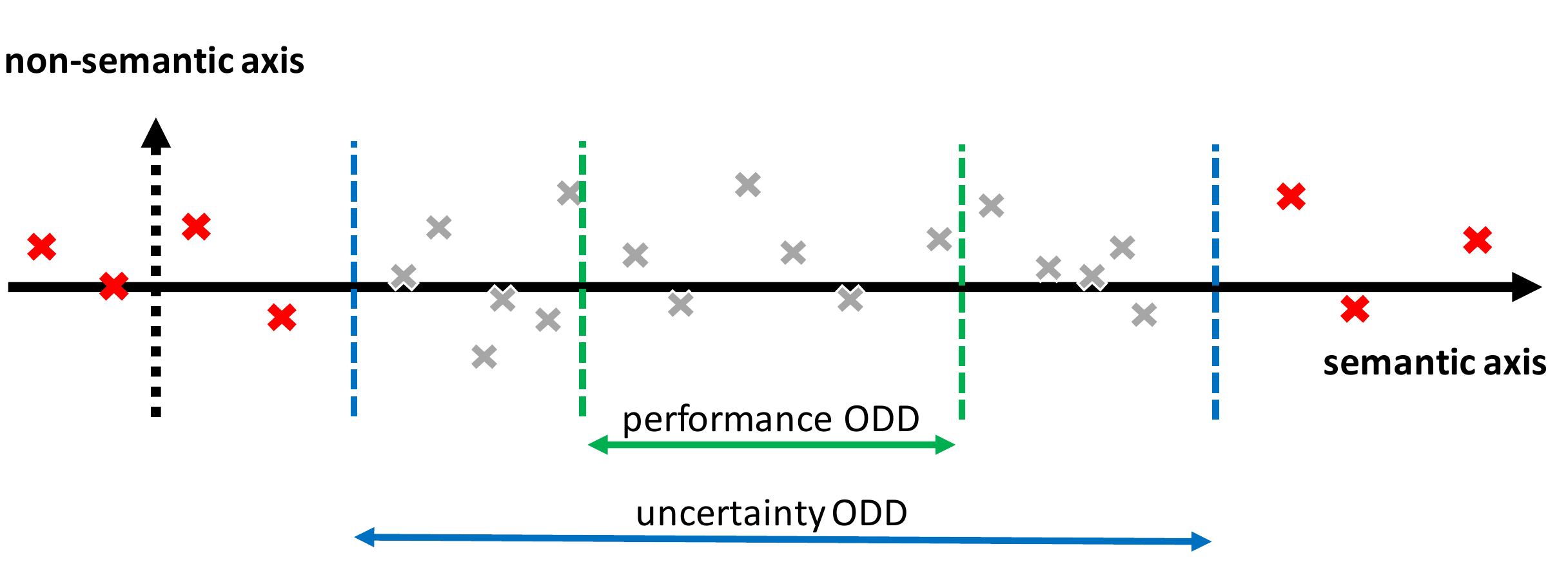}
    \caption{Schematic illustration of the \textit{ODD of an ML model} (green interval) and the \textit{ODD of its uncertainty estimator} (blue interval) in an input space that is sketched as a 2D plane. While the x-axis is considered to be semantically interpretable, the y-axis is non-semantic and thus not specifiable.
    The chosen semantic boundaries of the uncertainty ODD are wider compared to the model ODD, \ie in this example it is supposed to function properly in a larger area of input space. Moreover, the complementary technique of compiling in-domain (gray crosses) and out-of-domain (red crosses) input points is in accordance with the chosen semantic borders in this toy illustration: no red crosses are within the uncertainty ODD and no gray crosses outside of it. The non-semantic y-coordinates of these input points cannot be specified and thus fluctuate randomly.}
    \label{fig:unc_odd}
\end{figure}

One can moreover build on the ODD specification of the underlying DNN if given. 
Typically, however, the uncertainty estimator is supposed to also function reliably outside the original (performance) ODD of the DNN (see Fig.~\ref{fig:unc_odd}). 
Put differently, the uncertainty ODD is often larger than the performance ODD which may still provide a reasonable ODD ``core''.
Both the performance ODD
and the areas of the uncertainty ODD not overlapping with it 
will be addressed in testing. For those non-overlapping areas of the uncertainty ODD, a focus of testing are estimates of \mbox{epistemic uncertainty.}


\subsection{Modeling uncertainties}
\label{subsec:modeling_unc}

Technical specifications 
not only impact, as outlined, the desired properties and the operational design domain of an uncertainty estimator. Here, we investigate how they may
constrain uncertainty modeling \eg when using third-party ML components.
Aside, we analyze how ML model chains determine the ``flows'' of uncertainty information through one of their models (see second ``paragraph'').
Lastly, output granularities of DL systems and uncertainty estimators are studied (third ``paragraph''), as well as trade-offs and technical modeling constraints (last ``paragraph'').


\paragraph{Depth of integration into the ML model} 
\label{subsec:depth_integration}

To intertwine an uncertainty estimator with an ML model, the access restrictions the ML model is subject to need to be known as they impact both uncertainty modeling and testing. Such restrictions typically originate either from the state of model development (see next paragraph) or from black-box components with (to us) unknown architectures and weights that are employed in the ML system (see paragraph after next).

\textit{State of model development} 
As the development of a model progresses from initial planning over engineering its architecture towards optimization and evaluation, a deep integration of an uncertainty mechanism into it becomes less likely:
for a model under development, uncertainty modeling considerations may affect central building blocks of a DNN like architecture or objective function.
An almost ``finalized'' 
ML model, on the contrary, that cannot be modified any longer, constrains uncertainty estimators to be pre- or post-processing units. 
Such a limitation may also result from requirements on backward compatibility that ask \eg to re-use previously employed \mbox{model components.}

\textit{Black-box components} 
Many (pre-made) ML models underlie access restrictions, \eg black-box models from third parties with (to us) unknown structure and parameter values.
For such models a negative correlation between the extent of the restrictions and the depth of integration of the uncertainty estimator can be expected. 
As for ``finalized'' ML models (see above), one may rely on pre- or post-processing mechanisms or, for a model in a model chain (see below), on bypassing the uncertainty estimates from preceding ML components.
In contrast to the state of model development, the use of black-box components with access constraints directly impacts testing as they exclude testing methodologies that require knowledge about the internal state of a DNN.


\paragraph{Flow of uncertainty information}   
\begin{figure}[t]
    \centering
    \includegraphics[trim=90 70 170 60, clip, width=0.85\textwidth]{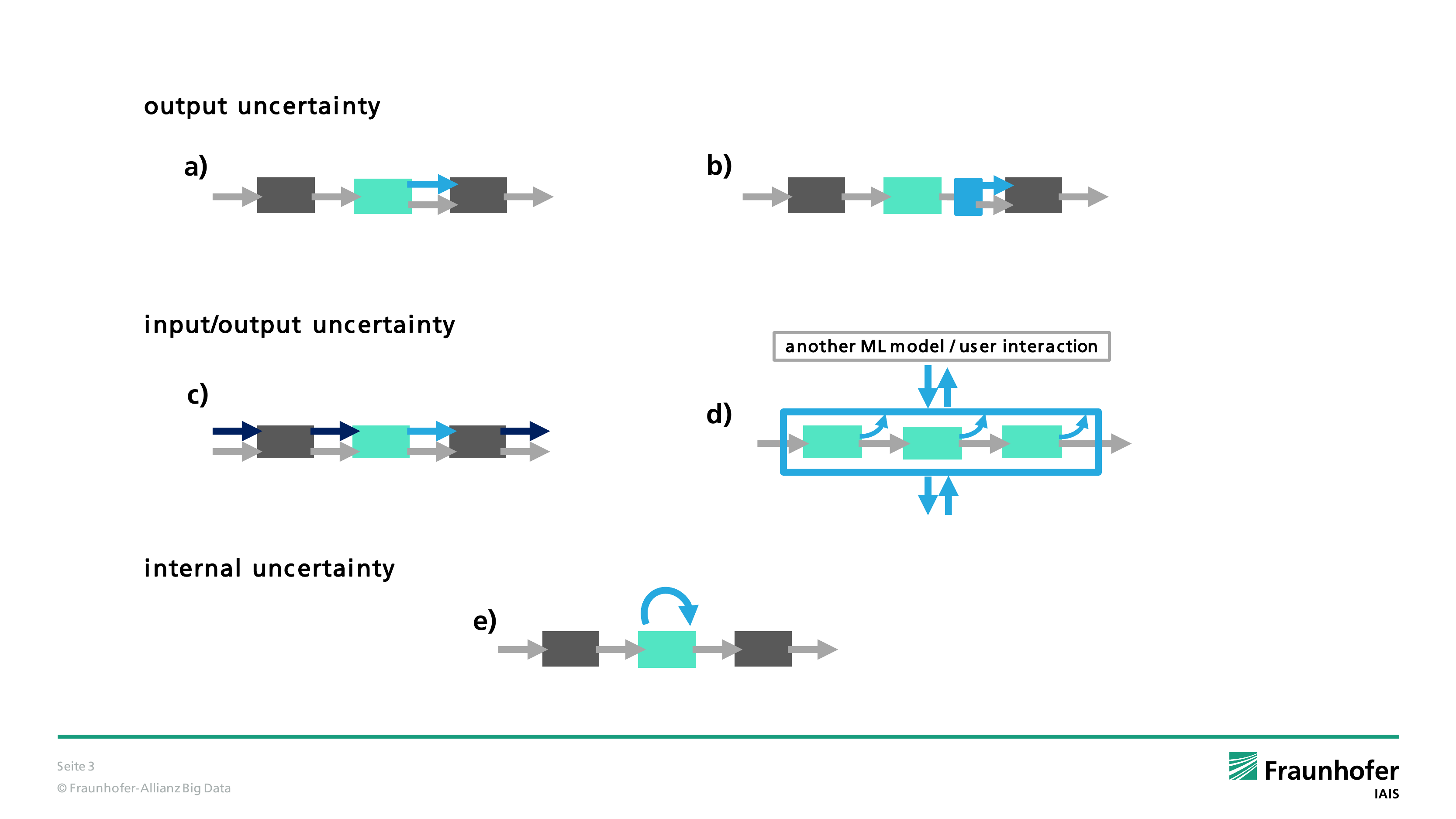}
    \caption{Prototypical ``flows'' of uncertainty information (blue arrows) along a chain of ML models (gray blocks and gray arrows). While uncertainty estimates for the entire chain are relevant (see d), we focus on uncertainty estimation for a single model in the chain (turquoise block, see a,b,c,e). In a) and b), output uncertainties are modeled and propagated to the subsequent model. The uncertainty estimator is integrated into the model (a) or a separate post-processing unit (b), respectively. An uncertainty estimator may incorporate propagated uncertainty information from preceding models (see c and d). Internal uncertainty estimates, on the contrary, are not propagated but generated and used within a model (see e) \eg to improve the quality of its output.
    }
    \label{fig:taxonomy_unc_flow}
\end{figure}
In many real-world ML systems, an ML model is not stand-alone but part of a model pipeline (see \eg \citet{sculley2015hidden}). Think \eg of a (prototypical) AD module stack that detects and characterizes vulnerable road users (VRUs), positions them in a 3D model of the recent traffic scene and predicts their future positions and velocities. Given the information flow along this model chain, various ways exist to enrich it with uncertainty information.
In autonomous driving, for instance, this information flow may be pedestrian proposals that are passed to a pedestrian classification network or, further down the AD stack, the 3D phase space coordinates of a detected pedestrian that serve as input for a motion prediction model. Adding uncertainty information, the VRU detections may be accompanied by confidence estimates and predicted future VRU positions (the mean estimates) may go along with estimated standard deviations. 
For such \textit{flows of uncertainty information} along a model chain, we provide a taxonomy (see Fig.~\ref{fig:taxonomy_unc_flow}) and discuss how the desired type of uncertainty flow impacts the development and testing of the \mbox{uncertainty estimator.}

We broadly distinguish between input, output and internally used uncertainties. 
Adding an uncertainty estimate to a model output
(see a) and b) in Fig.~\ref{fig:taxonomy_unc_flow})
is a standard scenario of uncertainty quantification. 
The two sketches in Fig.~\ref{fig:taxonomy_unc_flow} differ by the estimator's depth of integration into the considered (turquoise) model (see discussion in the previous ``paragraph''). While the estimator in a) is an integral part of the model, the estimator in b) is a post-processor, \ie a calculation routine which ``extracts'' an uncertainty estimate from the model's standard outputs. It may, for instance, calculate the entropy of a classifier's softmax output that poses a simple measure of inter-class uncertainty. Alternatively, an already existing uncertainty flow could be re-calibrated, using routines like temperature or Platt scaling \citep{guo2017calibration}.

Incorporating input uncertainties into a model's uncertainty estimation is crucial for many practical applications as such a propagation of uncertainty allows to eventually evaluate the behavior of an ML application as a whole, assuming a sufficiently high quality of the uncertainty estimates.
Research on the propagation of uncertainties \citep{wright2000input,ji2019propagation} has to deal with the limited availability of real-world datasets containing uncertainty-enhanced input features and the fact that such chains of ML models are highly application-specific, rendering them no typical object of study for a broad scientific community. 
For information-restricted black-box models, uncertainty propagation can be realized by bypassing uncertainty information, \ie by circumventing the considered ML model and instead piping the uncertainty information directly from the previous model to the subsequent one in the chain. 
A conservative uncertainty margin for the black-box model might be added.
Instead of modeling only one ``segment'' of this uncertainty flow (Fig.~\ref{fig:taxonomy_unc_flow}, c), some use cases require to ``zoom out'' and consider the entire model chain (Fig.~\ref{fig:taxonomy_unc_flow}, d). A critical capability of such a chain could be to incorporate external uncertainty information from surrounding software systems \eg to flexibly react to early indicators of a changing environment (such as a confusing traffic situation ahead). In the remainder of this work, however, we focus on one ML model in such a model chain.

Finally, some ML models employ uncertainty estimates internally, \ie neither the model inputs nor its outputs carry uncertainty information and thus there is no uncertainty information flow along the chain. 
The uncertainty estimates are instead generated and processed within the ML model. 
In such cases, uncertainty information is typically used to improve the model's performance, either directly for indirectly, \eg as a regularizer. 
\cite{henaff2019model} perform uncertainty-based regularization for trajectory prediction. 
\cite{he2019bounding} construct a two-stage object detection network whose region-proposal network uses a so-called var-voting scheme, \ie final detections are obtained by location-confidence-based aggregation of neighboring detections. Preliminary detections with low uncertainty thus have a higher impact on the final detection proposals.

Widening the focus, the flow of uncertainty information moreover influences quantitative testing (see sections~\ref{sec:setup_unc_testing} and \ref{sec:tests}).
Thoroughly evaluating, for instance, a setup like Fig.~\ref{fig:taxonomy_unc_flow} c) requires the availability of realistic input uncertainty data, while a setup like Fig.~\ref{fig:taxonomy_unc_flow} e) may have a focus on performance testing, rendering aspects like uncertainty calibration less important compared to most other uncertainty use cases.

\paragraph{Uncertainty granularity}

Given the types of uncertainty propagation presented above, we detail on the kind of information that flows and its dynamics.
Specifically, we compare the \textit{operation modes} and the \textit{output granularities} of the DNN and its uncertainty mechanism as they may diverge.
For an image-processing DNN, for instance, that operates frame-wise and at a constant frame rate, the uncertainty estimator may generate estimates at another frame rate, change-based or solely for inputs that are structurally different from its training data. 
This holds true for output granularity where the desired granularity of uncertainty information could be higher or lower compared to the one of the DNN outputs. While, for instance, a segmentation network generates outputs on pixel level, uncertainty information might be required on the level of objects, image parts or entire images. 
Contrarily, one may think of a pedestrian detection network to be equipped with uncertainty estimates on body-part level.
Such real-world ML systems often process structured information \ie both model inputs and outputs contain intrinsic (\eg spatio-temporal) dependencies, think \eg of image-processing DNNs that predict scene graphs or semantic segmentations.
Uncertainty modeling ought to reflect these dependencies, \ie ideally generates uncertainty estimates that \eg change steadily along a semantic segment, as opposed to estimates that strongly fluctuate from pixel to pixel.

\paragraph{Trade-offs and technical constraints}
Requirements are often at odds with one another, \eg the functional and the uncertainty-related requirements for an ML model.
In most cases, uncertainty requirements must not be met at the expense of considerably decreased functional quality and at most small deteriorations are tolerable. 
Moreover, we might face trade-offs between uncertainty estimators and other safe-ML techniques (that foster \eg interpretability or fairness) as all of them share the same neural capacities. 
Further technical constraints for modeling uncertainties range from system complexity over storage to latency and vary strongly from use case to use case.
Boundary conditions regarding time and resources not only apply to the uncertainty estimator itself but moreover to its development process where a simple ``off-the-shelf'' mechanism on the one hand and a scientifically novel custom estimator on the other hand mark extreme positions.


\secsummary

The technical boundaries of the last section together with the above discussed aspects of intended functionality
pose the high-level requirements the uncertainty estimator is subject to. 
All of them can be collected, either in an unstructured way or more formalized, using a requirement specification language (see \eg \citet{reinkemeier2011requirement}).
While these requirements reflect the initial demand for uncertainty estimation (see Fig.~\ref{fig:process_diagram}), they reside on various levels of abstraction and thus require further concretization to base the development and testing of uncertainty mechanisms on them.
In contrast, the operational design domain of the uncertainty estimator is already fully specified, \ie relevant semantic dimensions have been determined and value ranges along these dimensions were set.


\section{Towards uncertainty acceptance criteria}
\label{sec:acceptance_criteria}

In this section, we take steps to derive formalized acceptance criteria from the above collected high-level requirements.
For application-size ML systems, these requirements are often complex and thus ``translate'' into a variety of acceptance criteria.
These ``translations'', like any modeling, go along with simplifications and assumptions that inevitably stress and neglect certain aspects of the original requirements. 
To reduce resulting ``blind spots'', modeling ideally involves experts from different technical and non-technical domains (\eg safety engineers, ML researchers and product owners). 
The diversity of the acceptance criteria is further increased by formulating them redundantly, \ie by different experts who work independently of each other (also compare the qualitative ``Swiss cheese'' model in risk management \citep{perneger2005swiss}).
While these practices may increase the completeness of the acceptance criteria, it is beneficial to explicitly analyze whether completely fulfilled acceptance criteria actually capture the gist of the underlying requirements or whether they can be ``hacked'' (compare reward ``hacking'' in reinforcement learning \citep{hadfield2017inversereward}).

We help to structure and sharpen uncertainty requirements by providing a categorization of desirable uncertainty properties in subsection~\ref{subsec:desiderata}. While selected qualitative requirements can already be considered as acceptance criteria, the quantitative ones should be further formalized. For those, we introduce a formal notation of acceptance criteria as 3-tuples consisting of (\textit{semantic data specification}, \textit{measure of quality}, \textit{threshold value}) in subsection~\ref{subsec:acceptancecriteria}.


\subsection{Categorizing uncertainty requirements}
\label{subsec:desiderata}

Despite being case-specific at the level of details, qualitative and quantitative requirements for uncertainty estimation can be broadly categorized. 
We identify 10 categories, the first three of which capture different aspects of uncertainty quality. 
Categories four and five target conceptual properties of uncertainty estimation while categories six to nine focus on technical characteristics. 
They are complemented by a residual category for application-specific requirements that do not fit elsewhere.
In the following, we detail these requirement categories and highlight the benefits of fulfilling them:

\textit{Aspects of uncertainty quality}
\begin{itemize}
    \item \textbf{calibration} 
    Uncertainty estimates are calibrated when their distribution matches the distribution of actual model errors \citep{degroot1983comparison,naeini2015calibrated}. 
    These matchings are typically calculated across an entire dataset, rendering calibration a dataset-wide (or global) property.
    Global calibration is quantified by measures like ECE and desirable for both in-data (ID) inputs, \ie inputs from within the ODD (or more concretely, from a dataset approximating it) and out-of-data (OOD)\footnote{Be aware that the abbreviations ODD and OOD bear the risk of confusion. 
    While ODD is the operational design domain, OOD stands for out-of-data, \ie they refer to diverging concepts that are (in a qualitative sense) even contrary to one another.} inputs.
    A small ECE \eg assures that uncertainty estimates statistically resemble the model errors. While it does not imply that each individual large uncertainty estimate goes along with a large model error and vice versa, it allows to assess the on-average quality of the uncertainty mechanism and provides a foundation for more detailed statistical analyses (see subsequent categories). 
    
    \item \textbf{local calibration / heteroscedasticity}
    Uncertainty estimates are \textit{locally} calibrated when they probabilistically match prediction confidence for each data point. 
    This allows local inference on the model's trustworthiness for a given input. 
    As most datasets do not provide ground truth uncertainties, local calibration is typically subject to resolution limits.\footnote{Assuming however a dataset that comes along with ground truth uncertainties $\sigma_{\textup{gt,i}}$, \ie data points with labels $(\mu_{\textup{gt,i}}, \sigma_{\textup{gt,i}})$, and a model with predictions $(\mu_{\textup{i}}, \sigma_{\textup{i}})$, the model's local calibration could be evaluated \eg by means of a Wasserstein distance,
    \ie as $\sum_i \left((\mu_{\textup{i}} - \mu_{\textup{gt,i}})^2 + (\sigma_{\textup{i}} - \sigma_{\textup{gt,i}})^2\right)$.}
    An assessable proxy measure is ECE calculated on (local) data subsets, \eg on a subset of safety-critical scenes. A high ECE on such inputs could reveal weaknesses of an uncertainty estimator that remained undiscovered when calculating the ECE averaged over a large dataset as \eg underestimated uncertainties for safety-critical inputs might have been balanced by overestimated uncertainties for not safety-critical inputs. While even input-independent, constant uncertainty estimates could lead to small global ECE values \citep{degroot1983comparison}, local calibration is closely linked to input-dependent heteroscedastic uncertainty estimates.
    Local calibration is desirable for ID data points and a variety of OOD scenarios. 
    It is a strict criterion that implies global calibration and the ability to solve downstream tasks (see next category).
    
    \item \textbf{ability to solve downstream tasks} 
    In case an uncertainty estimator is used as an active component in an ML system, its estimates are often used as input for downstream decision making, especially whether safe and reliable operations of the system can be ensured on the current inputs. The choice of the concrete downstream task to solve is highly application-specific and may, for instance, be the detection of inputs that are structurally different from training data (ID-OOD detector) to enable, \eg for AD systems, safe stopping if necessary. Another example from automated perception is the recognition of false detections (\eg of VRU reflections in windows) which are then no longer considered \eg for motion planning. 
    In an industrial context, automated inspections that go along with high uncertainty estimates might trigger manual re-evaluations of individual work pieces.
    Technically, sufficiently strong correlations between uncertainty estimates and actual model errors are required to enable this kind of ``self-assessment''.
    Compared to calibration (see above), less attention is paid to specific distributional properties of the uncertainty estimates. 
    Instead, they are employed as a tool that often remains useful even when some desirable distributional properties are not met. 
\end{itemize}
\textit{Conceptual properties}
\begin{itemize}
    \item \textbf{argumentatively substantiated} 
    Broad scientific analyses, both theoretical and empirical, help to draw a realistic picture of the strengths and weaknesses of an uncertainty mechanism. 
    On the one hand, theoretical justification allows for a more thorough understanding of the mechanism and thus enables more confident predictions of its behavior (\eg on previously unseen data). In the long run, these investigations aim for theoretical verifications of DNN-based estimates (see \eg \citet{salman2019convexrelaxation}).
    A large number of empirical analyses, on the other hand, render unknown structural weaknesses of a mechanism less likely. 
    Thus, for critical use cases, algorithms are favorable that are both theoretically and empirically substantiated.
 
    \item \textbf{attribution to type of uncertainty} 
    Better understanding what causes (unacceptably) high uncertainty estimates 
    allows to take appropriate countermeasures. Attribution is typically given \wrt either epistemic or aleatoric uncertainty, see related work in section~\ref{sec:related}. The former
    corresponds to model-weight uncertainty and may be reduced by adding 
    further
    input scenarios to the training data. 
    The latter one, data-intrinsic uncertainty, is more difficult to handle and may require changes to the data collection process or the modeling task.  
    For many real-world uncertainty use cases, clear distinctions between epistemic and aleatoric uncertainty are difficult since their training datasets are very sparse samples from high-dimensional input spaces.
\end{itemize}

\textit{Technical characteristics}
\begin{itemize}

    \item \textbf{applicability to large neural networks} 
    Most real-world ML systems employ DNNs with millions of parameters. Applicability to, in that sense, large networks might thus be a desirable practical prerequisite for uncertainty estimation techniques.
    
    \item \textbf{minimal overhead} 
    Calculating uncertainty estimates often requires additional numerical operations (\eg for sampling-based approaches) or extensions to the neural model (\eg for ensembling-based techniques). 
    As many applications underlie resource constraints (especially mobile systems such as phones or drones) and more generally for sustainability reasons, it is favorable to keep the ``foot prints'' of uncertainty estimators minimal \wrt storage and computing. 
    
    \item \textbf{minimal trade-offs} 
    While uncertainty quality and model performance are (in most cases) orthogonal requirements, there may be trade-offs between them in practice.
    This can be seen, for instance, on mechanisms that share weights to perform both task- and uncertainty-related predictions.
    It might therefore be desirable to use an uncertainty mechanism that leads to only minimal deteriorations in performance or other safety-relevant metrics.
    
    \item \textbf{technical simplicity} 
    Being technically simple facilitates the integration of an uncertainty mechanism into 
    neural architectures.
    Estimators, on the contrary, that require substantial changes to a network like additional layers or that rely on specifically annotated training data are harder to implement. They therefore incur a larger technical debt, which might make maintenance more challenging and increase the risk of programming errors.
\end{itemize}

\textit{Application-specific requirements}
\begin{itemize}
    \item
    Depending on the use case various additional requirements may exist.
    Semantic segmentation tasks, for instance, generate highly granular uncertainty outputs that render the following two requirements plausible: first, one may ask for coherent uncertainty estimation for the pixels belonging to a single object in an image, and second, one may expect pixel classification uncertainty to increase when approaching a ``boundary" between semantic segments as such boundaries go along with irreducible data uncertainty. For the related computer vision task of object detection, it seems natural to ask for a positive correlation between estimated uncertainty and the degree of occlusion of a VRU. In case of combined segmentation and object detection, one could require high segmentation uncertainties for image areas that contain undetected objects (false negatives).
\end{itemize}

This categorization scheme allows to more clearly delineate the initial requirements, \eg by splitting or merging them. It might moreover help to identify additional requirements that extend the initial set.
While some of these requirement categories are qualitative, \eg ``being argumentatively substantiated'', others are quantifiable, \eg ``being globally calibrated''. 
The (relative) importance of the different requirement categories is moreover not fixed but varies for each uncertainty use case.
In the following, we focus on the quantifiable uncertainty requirements and take steps to derive acceptance criteria from them.


\subsection{Formalizing uncertainty acceptance criteria}
\label{subsec:acceptancecriteria}

Having disentangled the textual requirements using the categorization above we further address those sub-parts that are specifically quantitative, \eg issues of calibration.
Concretely, we ``translate'' each quantitative requirement into an acceptance criterion that consists of three key elements: first, a \textit{semantic data specification} that reflects those regions of the input space that are relevant for a given requirement. Second, \textit{a measure of quality}
that quantifies to which degree a requirement is fulfilled and, lastly, a \textit{threshold value} that determines whether this degree of fulfillment is sufficient for the given DL system.
While for some requirements, \eg ``minimal overhead'', measures of quality are not uncertainty-related, \eg run time, we focus on uncertainty-related aspects in the subsequent exposition.
Please note that one uncertainty requirement may map onto several acceptance criteria.

\paragraph{Semantic data specification}
The semantic data specification of an acceptance criterion can be obtained based on the semantic dimensions that were used to describe the uncertainty ODD (see subsection \ref{subsec:modal_and_domain}). 
For each of these semantic attributes, it is checked whether the uncertainty requirement renders it necessary to focus on certain value ranges and to exclude others. 
Requiring, for instance, that an autonomous car overtakes a cyclist only if it is highly certain that a safety distance of at least 1m can be kept at any point of this maneuver, might translate into a data specification that not only requires at least one cyclist per image but moreover a focus on short distances between cyclist and vehicle.
The traffic participants aside, the road type of the input scenario is to be specified:
while highways and country roads are not irrelevant, a clear focus for overtaking maneuvers is on urban environments.
In a subsequent step (see section~\ref{sec:setup_unc_testing}), the accordingly specified regions of the input space are mapped onto concrete datasets.
In the following, we focus on the two other components of uncertainty acceptance criteria and provide best practices for choosing uncertainty measures (next subsection) and their respective threshold values (subsection after next).


\paragraph{Measures of uncertainty quality} 

To determine to which degree requirements like global and local calibration are fulfilled and to quantify potential trade-offs between model performance and uncertainty estimation, scores of uncertainty quality are needed.
Mathematically, these scores aim at concise descriptions of the mismatches between (often) high-dimensional probability distributions, mostly by means of scalar scores. 
These reductions of dimensionality bring about different sensitivities and distortions, compare \eg 2D map projections of a 3D globe where no map preserves areas, shapes and distances at the same time.
Considering multiple uncertainty metrics, however, allows to ``construct'' a more complete picture. This is of special relevance when optimizing a model on an uncertainty-related measure like negative log-likelihood (NLL, see section~\ref{sec:related}). The resulting NLL-optimized model may neglect aspects of uncertainty that are not captured by NLL and alternative measures like ECE
can be used
to detect such overfitting to the optimization metric (in some fields also known as Goodhart's law \citep{chrystal2003goodhart}).

Aspects to differentiate between uncertainty measures are (among others) their sensitivity to underestimated and overestimated uncertainties as well as their sensitivity to outliers (in both directions). 
NLL, for instance, is unbounded and thus more sensitive to outliers compared to the bounded ECE. 
While this insensitivity of ECE to strongly deviating uncertainty estimates limits its ability to resolve quality differences for a broad range of models, such outlier suppression might be acceptable for analyses that aim at measuring on-average uncertainty quality.
Outliers aside, NLL is more sensitive to under-estimated uncertainties (asymptotically $\propto 1/\sigma^{2}$) compared to over-estimated ones $(\propto \log(\sigma))$.
This implies that NLL-optimized models (on average) tend to overestimate uncertainties; a conservative behavior that might be considered advantageous from a safety perspective.
ECE on the contrary treats over- and underestimated uncertainties roughly equivalent. 

Taking a testing perspective, a point-wise measure like NLL has favorable properties compared to set-based measures like ECE. Search-based test (SBT) strategies explore the input data space to detect \eg input regions for which the model's uncertainty quality is low. 
For NLL, each new input data point provides a feedback on the (local) ``success'' of the search strategy as the point can be compared to its predecessors along the search trajectory. ECE in contrast requires a local data sample or the stored recent history of the search trajectory.
However, NLL is not a calibration measure and thus not suited for the related requirement categories, namely global and local calibration. 

Both NLL and ECE capture on-average properties of uncertainty estimators as opposed to scores that ``zoom in'' and put emphasize on specific (safety-relevant) quantile ranges of uncertainty distributions. The expected tail loss (ETL, see section~\ref{sec:related}), for instance, measures the ``depth'' of \mbox{distributional tails.}

Besides these generally applicable scores, task-specific uncertainty measures like probability-based detection quality (PDQ, \cite{hall2020probabilistic}) for object detection 
and area under the sparsification error curve (AUSE, \cite{ilg2018uncertainty}) for optical flow estimation exist. Reflecting the respective structures of inputs and outputs, they capture task-specific aspects of uncertainty and thus complement the generally applicable measures. 
For complex uncertainty use cases, one may additionally construct custom uncertainty measures, \eg to involve prior knowledge about image segments or object types.

Finally, the standard versions of most uncertainty scores assume Gaussianity for model errors and uncertainty estimates, \eg NLL and ECE. While these assumptions are (partially) justified for some modeling techniques or technical limits, empirical testing is called to analyze their validity for a given DL system.

\paragraph{Thresholding uncertainty measures}

To get from quantitative uncertainty measures to uncertainty tests that are typically formulated as binary decision rules, we require threshold values that indicate whether a test succeeds or fails.
This binarization ``transforms'' statistical evaluations, that ask how good the quality of an estimator is, into semantic evaluations, asking whether the quality of an estimator is good enough for an application context.
The following edge cases illustrate how strongly the notion of a semantically appropriate threshold value varies from use case to use case. 
For gambling, on the one hand, a win rate of $50.1\%$ (for a binary game) may suffice to generate a profit \textit{on average}, assuming the absence of fees and unlimited scalability.
Critical ML applications, on the other hand, could require safe operations in \textit{each individual situation}. In this case even well-functioning in $99.9\%$ of all scenarios could 
indicate that in $1$ out of $1000$ situations a subsequent safety strategy should be considered.

Moreover, the already discussed challenges for breaking down high-level requirements become especially apparent for threshold values as these are typically formulated on a system level. To derive, for instance, a threshold value for an object detection (OD) network from a high-level requirement such as a maximum tolerable VRU collision rate, this threshold value, the collision rate, 
can be propagated
along the AD model stack to the considered OD network, an inevitably qualitative process that requires various additional assumptions.

For a given DL system, an uncertainty metric may go along with several threshold values as the latter ones are set on the level of acceptance criteria and thus depend on their data specifications, \eg whether the entire ODD, parts of it or transitions towards out-of-data are considered (see section~\ref{sec:setup_unc_testing} for discussions on test datasets). 
Setting a threshold value for the negative log-likelihood (NLL) is especially challenging as it is a hybrid measure 
of network performance and uncertainty quality. 
A threshold value thus has to reflect both performance- and \mbox{uncertainty-related requirements.}


\secsummary

The threshold values together with the above discussed uncertainty measures and semantic data specifications compose the quantitative acceptance criteria. Each quantitative requirement (from section~\ref{sec:initial_demand}) should be formalized accordingly. Please recall that while uncertainty measures and threshold values are already fixed, the data specifications are still qualitative. Test datasets will be derived from them at the level of test cases in section~\ref{sec:tests}. These quantitative acceptance criteria, together with the qualitative ones, provide the
``gold standard'' for the next steps of the framework: 
first, the actual selection and adaptation of an uncertainty mechanism (step~3) and second, its quantitative testing (steps~4 and 5, compare respective arrows in Fig.~\ref{fig:process_diagram}).


\section{Choosing an uncertainty mechanism}
\label{sec:mechanism}

Comparing the desirable uncertainty properties outlined in the previous section with the actual properties of modern uncertainty modeling techniques, allows to choose suitable uncertainty estimators.
To this end, we mirror the structure of subsection \ref{subsec:desiderata}
and analyze if and how well the 10 qualitative and quantitative requirement categories are addressed by representative uncertainty mechanisms. 

\textit{Aspects of uncertainty quality}
\begin{itemize}
    \item \textbf{calibration}
    Self-assessing model performance 
    involves comprehensive uncertainty quantification, \ie modeling
    both aleatoric and epistemic uncertainty. 
    For transitions from ID to OOD, the importance of epistemic uncertainty (relative to aleatoric uncertainty) increases, rendering its modeling
    important
    for open-world uncertainty use cases.
    While post-calibration routines, that often adjust a small number of global model parameters, help to improve uncertainty calibration in-distribution, this benefit does not necessarily carry over to out-of-distribution data (see \eg \cite{snoek2019can}).
    More advanced methods like MC dropout applied to networks that output the parameters of a Gaussian distribution \citep{kendall2017uncertainties} and deep ensembles \citep{lakshminarayanan2017simple} improve on both ID and OOD calibration by combining parametric approaches that are optimized to capture ID aleatoric uncertainty with matching-based strategies for modeling \mbox{epistemic uncertainty.}
  
    \item \textbf{local calibration / heteroscedasticity}
    Aleatoric uncertainties are typically heteroscedastic, \ie the data-intrinsic noise varies from datapoint to datapoint.
    While a homoscedastic modeling of heteroscedastic aleatoric uncertainty (like MC dropout) allows to grasp the average level of data noise, it is more common to model it more flexibly \eg by means of network outputs that are interpreted as the parameters of a Gaussian distribution (as is done by an extension of MC dropout due to \cite{kendall2017uncertainties}).
    Such parametric approaches aside, non-parametric approaches to modeling heteroscedastic aleatoric uncertainty exist, such as the second-moment loss \citep{sicking2021sml} and Wasserstein dropout \citep{sicking2021wdrop} that reflect the datapoint-dependent noise in the width of the network's output distribution that is generated by sampling sub-networks.
    While model-inherent (epistemic) uncertainty is in general also heteroscedastic, it typically increases, unlike aleatoric uncertainty, when leaving the training data distribution. The above outlined (implicit) ensembling methods model epistemic uncertainty by the ``spread'' of their sub-networks.  On training data, these (sub-)networks are ``matched'', resulting in a small ``spread''. Under data shifts, these ``matchings'' do not hold, the (sub-)networks ``disperse'' and thus yield larger uncertainty estimates.
    
    \item \textbf{ability to solve downstream tasks}
    While the evaluation of uncertainty estimators by means of downstream tasks is common in research, the exact tasks considered vary from method to method, rendering many results not readily comparable.
    Common types of tasks are the detections of misclassifications and outliers as well as distinctions between in- and out-of-distribution data. 
    The latter task is specifically suited for uncertainty as it touches on the problem of uncertainty attribution (see respective category).
    Some methods explicitly learn to distinguish between different types of uncertainty (see \eg \citet{malinin2018predictive}) and thus simplify downstream OOD detection.
    However, approaches in that spirit work only for low-dimensional data as, in many cases, explicit OOD data is needed.
    In high dimensions, comparisons between ID and OOD are often realized by evaluations on structurally different datasets (see \eg \citet{snoek2019can}).
    These analyses pose rather rough approximations of the continuous data shifts typically encountered in the real world.
\end{itemize}

\textit{Conceptual properties}

\begin{itemize}
    \item \textbf{argumentatively substantiated} 
    Many uncertainty estimation techniques (\eg stochastic gradient Langevin dynamics \citep{welling2011bayesian}, MC dropout \citep{gal2016dropout}, deep evidential regression \citep{amini2020evidential}) have their roots in Bayesian statistics and are, to varying degree, motivated in this regard, giving them a theoretical basis.
    Other techniques like deep ensembles take a frequentist perspective. 
    Deep ensembles as well as MC dropout are considered to be among the most established uncertainty modeling techniques when judged by the high numbers of related publications and \mbox{open-source implementations.} 
    
    \item \textbf{attribution to type of uncertainty}
    Attributing an uncertainty estimate to uncertainty sources requires a mechanism that models several types of uncertainty in the first place, especially aleatoric and epistemic uncertainty.
    As stressed in the related work, most such mechanisms internally employ combinations of ``atomic'' uncertainty mechanisms each of which modeling one type of uncertainty.
    Attributing uncertainty comes for such approaches ``free of charge'', \eg for deep ensembles or MC dropout applied to networks that output the parameters of a learned Gaussian distribution.
    A fully parametric approach that provides a comparably easy decomposition is deep evidential regression. 
    Prior networks for classification \citep{malinin2018predictive} take uncertainty-source attribution one step further as they allow to moreover distinguish distributional uncertainty from data and model uncertainty.
\end{itemize}

\textit{Technical characteristics}
\begin{itemize}
    \item \textbf{applicability to large neural networks} 
    The methods discussed in the related work (section~\ref{sec:related}) were pre-selected having a sufficient scaling behavior to be employed for application-size ML systems. Ensembling- and subsampling-based mechanisms, for instance, scale linearly with model size. 
    For use cases employing small DNNs, the scope of practically applicable uncertainty mechanisms broadens as various Bayesian neural network (BNN) approximations (\eg \cite{liu2019accurate}) become affordable.
    
    \item \textbf{minimal overhead}
    We analyze the overheads of uncertainty mechanisms \wrt numerical operations (processing time) and storage. Depending on the context, further ``dimensions'' may be considered such as the transmitted data volume in federated-learning applications (see \eg \citet{mcmahan2017federated,kamp2018federated}).
    Regarding numerical operations, parametric methods \citep{nix1994estimating} cause virtually no overhead as opposed to subsampling- and ensembling-based methods whose overhead is (typically) linear in the number of subsamples and model components, respectively. 
    Similar considerations hold for networks like probabilistic U-net \citep{kohl2018probabilistic} that introduces an additional network branch and relies on sampling in a latent space.
    Efficiency-optimized versions of subsampling mechanisms such as last-layer variants \citep{snoek2019can} and approximate analytical moment propagation \citep{postels2019sampling} allow for noteworthy reductions of the numerical operations and thus processing time.
    Regarding storage, (non-weight-sharing) ensemble methods (and approaches in that spirit that add \eg a network branch \citep{kohl2018probabilistic}) require additional resources. 
    Subsampling approaches (\eg MC dropout or masksembles), \ie in a sense weight-sharing ensembles, do not introduce any additional storage overhead.
    
    \item \textbf{minimal trade-offs}
    Uncertainty mechanisms impact network performance, in most cases, by modifying either the network capacity or its optimization objective.
    A parametric model for a regression task, for instance, does the latter as it optimizes Gaussian likelihoods (NLL) instead of squared errors (RMSE). 
    This simultaneous optimization of mean and variance typically causes a certain decrease of performance (at least on training data). 
    On the contrary, both subsampling- and ensembling-based methods change the network capacity, however, in opposite directions: 
    subsampling limits the capacity as for each forward pass weights are omitted while ensembling increases the capacity as multiple models are learned for the same task.
    Trade-offs (or synergies) between uncertainty estimators and other trustworthy-ML components can be expected (see section \ref{sec:related}), but are difficult to quantify due to the large variety of the latter ones. 

    \item \textbf{technical simplicity}
    Being technically simple is a driving factor for the wide adoption of an uncertainty mechanism.
    The modifications on code level induced by standard approaches such as subsampling-based, ensembling-based and parametric ones are in most cases minimal. For the mentioned approaches, the network is slightly modified (subsampling), a loop over model training is added (ensembling) and the objective function is modified (parametric), respectively. 
    Examples for more involved techniques are SWAG \citep{maddox2019simple} and probabilistic U-net \citep{kohl2018probabilistic} that aggregate statistics along the training trajectory and require an additional network branch, respectively.
\end{itemize}

\textit{Application-specific ones}
\begin{itemize}
    \item The main types of uncertainty estimators are flexibly enough to be employed in a variety of neural architectures and allow to estimate uncertainties at different granularity levels.
    Of special importance are low-level uncertainty information \eg on pixel level, as they allow to construct custom uncertainty aggregates for ML applications that require non-standard uncertainty ``entities''. Approaches like probabilistic U-net (ibid.) help to strengthen the coherence between such pixel-level uncertainty estimates. To further improve the susceptibility of a segmentation model \wrt critical ``segments'' such as VRUs, one may weight the segmentation outputs with pixel-resolved prior class probabilities (see \eg \cite{chan2019decisionrules}).
    
\end{itemize}

Many state-of-the-art uncertainty estimators combine different types of mechanisms to mitigate conceptual weaknesses and to leverage their respective strengths, \eg to describe both aleatoric and epistemic uncertainty. 
Complex use cases may require an adaption of existing mechanism combinations or even the construction of new ones. 
One may further analyze whether an uncertainty mechanism introduces (steerable) trade-off relations between some of the 10 requirement categories. 

\secsummary

Having picked an uncertainty mechanism, it 
should
be optimized. 
Some mechanisms render post-training of the ML model or even re-training from scratch necessary, for others a less involved post-processing routine suffices (see \eg temperature scaling \citep{guo2017calibration}). 
The ``original'' network without uncertainty estimator may be kept for comparisons \eg to detect ``side effects'' of integrating the uncertainty mechanism such as potentially occurring reduced task performance. 
    

\section{Scope and structure of uncertainty testing}
\label{sec:setup_unc_testing}

Given an optimized uncertainty estimator, we are headed to test whether it fulfills the uncertainty acceptance criteria (see section~\ref{sec:acceptance_criteria}). 
Before we construct concrete test cases in section~\ref{sec:tests}, we introduce a test hierarchy that builds on semantic and non-semantic test data selection strategies and discuss how it aids in 
the testing of the uncertainty acceptance criteria.
We focus on quantitative testing as qualitative acceptance criteria (\eg being theoretically justifiable or allowing for uncertainty attribution) were (in most cases) already addressed when choosing an uncertainty mechanism in the previous section. The goal of this step in the framework (recall Fig.~\ref{fig:process_diagram}) is to determine the test ``depth'' and the test focuses for each quantitative acceptance criterion.

Different from testing programmed software, approaches that cut a DNN in smaller pieces are challenging due to the highly interacting non-linear (black-box) nature of these models, also referred to as ``changing anything changes everything" (CACE, see \cite{sculley2015hidden}).
We therefore consider the entire DNN as a single unit for testing.
Thus, test hierarchies from classical software testing (from unit over integration to system tests) are not easily transferable (see \eg \citet{gannamaneni2021concepttesting}).
The complexity of (real-world) ODDs moreover renders comprehensive testing of (uncertainty) acceptance criteria 
difficult
(see outline in section~\ref{sec:intro}). 
Diverse testing types and testing methods however yield a broad overview of the strengths and weaknesses
of an uncertainty estimator within a given computing budget.

\begin{figure}[t]
    \centering
    \includegraphics[trim=25 110 15 60, clip, width=1.00\textwidth]{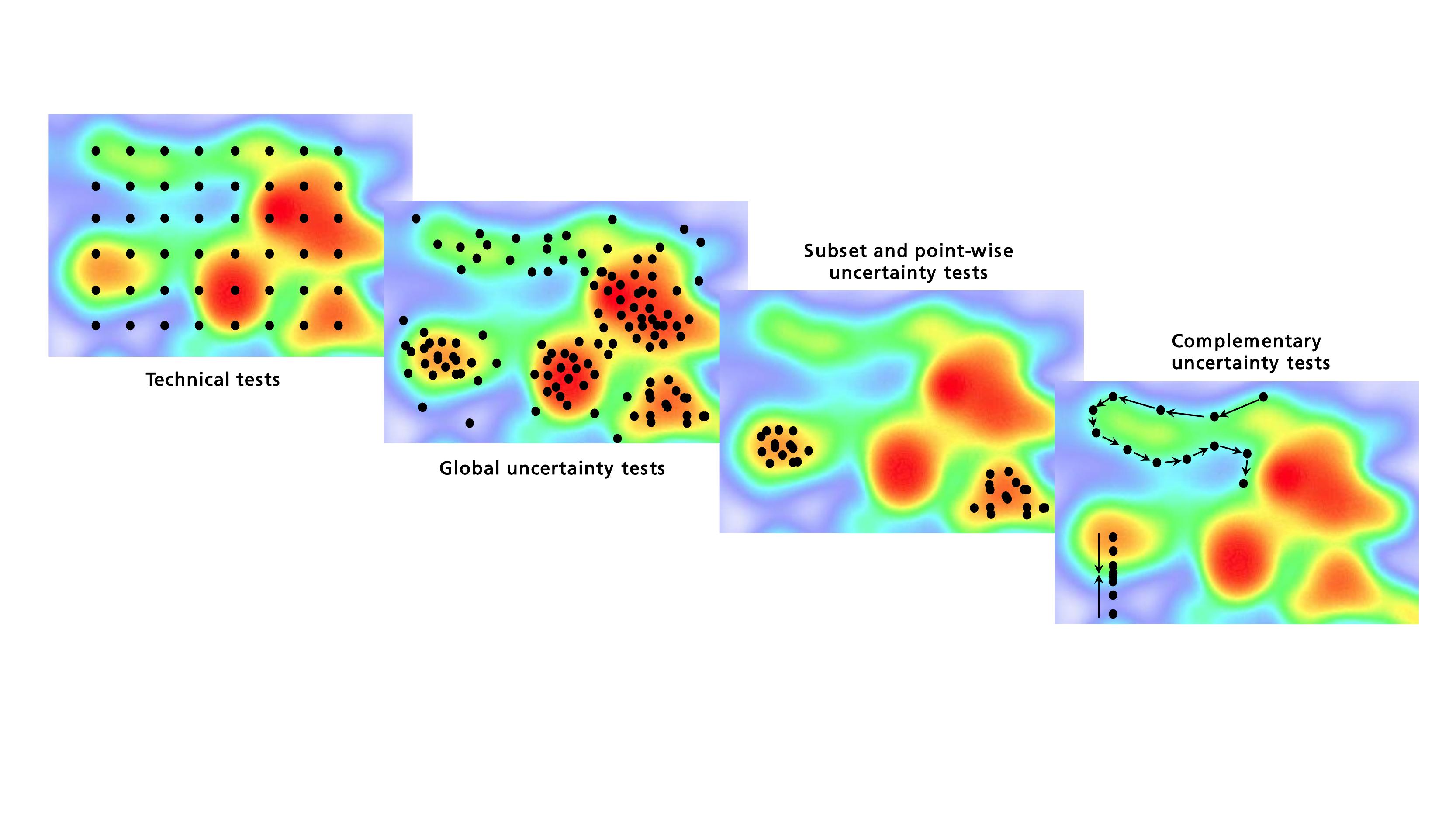}        
    \caption{Schematic visualization of the data selection concepts on the different levels of the test hierarchy. 
    The heatmap shows a fictitious data density (violet is low, red is high).
    Simple technical tests (left panel) are based on a coarse data selection. Global uncertainty tests in contrast (second panel), rely on a broad, density-appropriate sampling of the input space. Subsequent subset and point-wise uncertainty testing (third panel) focuses on particular input scenarios and regions. Complementary uncertainty tests (right panel) finally evaluate more complex properties of uncertainty estimates.
    The actually conducted tests strongly depend on the uncertainty use case, see the following subsections for details.
    }
    \label{fig:test_hierarchy}
\end{figure}

We propose a test hierarchy that builds on iteratively refined test data selection strategies,
as it is mostly not possible to conclusively map the ODD (or representative parts of it) onto a dataset.
The data selection hierarchy is a practical vehicle serving a twofold purpose: on the one hand, it allows to broadly cover the entire ODD and, on the other hand, to conduct deep-dive analyses exploring challenging and \mbox{(safety-)critical} regions of it.
Specifically, we distinguish four hierarchy levels:
\begin{itemize}
    \item \textit{Technical tests} (subsection \ref{subsec:tech_tests}) that check elementary technical properties of an uncertainty estimator. The semantics of the employed test inputs are (largely) irrelevant.
    \item \textit{Global uncertainty tests} (subsection \ref{subsec:global_tests}) that focus on global properties of an uncertainty estimator such as its on-average quality within and outside the performance ODD. 
    \item \textit{Subset and point-wise uncertainty tests} (subsection \ref{subsec:subset_tests}) analyze worst-case behaviors (and other quantile ranges) of the uncertainty estimate distribution and how well an uncertainty estimator functions for individual (application-critical) input scenarios. 
    \item \textit{Complementary uncertainty tests} (subsection \ref{subsec:complementary_tests}) are an open residual test set to explore more involved uncertainty properties as well as novel testing methods.
\end{itemize}
For a schematic visualization of the test data selection concepts on the four hierarchy levels, see \mbox{Fig.~\ref{fig:test_hierarchy}}. Before introducing the test hierarchy in detail, we stress that it is applied to each (quantitative) uncertainty acceptance criterion separately:
first, we determine the \textit{test depth} for the criterion, \ie the levels of the test hierarchy to be considered for adequately addressing it. 
Each of these test hierarchy levels in turn contains several test types, each of which putting a focus on another relevant characteristic of uncertainty estimates. 
These test types are finally broken down onto concrete test cases (see section~\ref{sec:tests}).

\subsection{Technical tests}
\label{subsec:tech_tests}

A first step towards thoroughly examined uncertainty estimates is testing their basic technical properties. 
We begin by checking the interfaces for uncertainty information, especially whether they comply with the flows and granularities determined in subsection \ref{subsec:modeling_unc}. 
The data types of the interfaces are relevant as they set the ranges and resolutions of the uncertainty estimates. 
Given the interfaces, we test how an estimator handles undefined or invalid inputs, \eg a corrupted negative uncertainty estimate from a previous ML model in a model chain and whether explicit sanity checks for uncertainty outputs are implemented, \eg whether a flag is raised for negative or close-to-zero estimates. 
For black-box ML models with access constraints, these output-related checks cannot be conducted as they require knowledge about the model's inner workings. 
Some uncertainty mechanisms (applicable to ``white-box'' networks) go along with additional model parameters (\eg ensembling methods, see section \ref{subsec:modeling_unc}).
Such mechanisms call for analyses of whether these additional parameters were properly adjusted by network (re-)training, \ie if their values are \eg continuously distributed or that none of them is undefined.

Running an uncertainty mechanism on a larger test dataset allows to examine its processing time distribution that provides insights on average processing times (especially relative to the processing times of the same DNN without an uncertainty estimator) and potentially occurring latency ``tails'' that need to be handled.
Finally, one may require uncertainty estimators to be deterministic for test reproducibility. 
As many uncertainty mechanisms rely on sampling routines, this requirement translates into the question of whether all employed pseudo-number generators are seeded.


\subsection{Global uncertainty tests}
\label{subsec:global_tests}

Recalling section \ref{subsec:modal_and_domain}, the uncertainty ODD is, in most cases, intended to be a superset of the DNN's performance ODD, \ie the uncertainty estimator is expected to reliably predict the model's confidence on inputs from both within (ID) and outside (OOD) the model's training distribution. 
Therefore, global uncertainty tests investigate the general functioning of an uncertainty estimator on the broader uncertainty ODD and are conducted on accordingly compiled datasets. 
Intermediate test results per data point are (typically) aggregated to provide dataset-wide uncertainty quality figures.

For testing within the performance ODD, one may build, if available, on the datasets used for the performance evaluation of the network or otherwise on structurally similar datasets.\footnote{In general, we assume that both network and uncertainty estimator are (concurrently) trained on the same dataset. Assuming further that the uncertainty estimator captures epistemic uncertainty, we expect its ``domain of proper functioning'' to be larger compared to the one of network performance.} 
Possibly occurring performance degradation due to the inclusion of an uncertainty estimator (performance-uncertainty trade-off) should be investigated. 
While further insights on performance may arise during the global tests, they are not the primary focus.

Next, we examine the uncertainty estimator's quality in the ``outer'' areas of the uncertainty ODD, \ie on data structurally different from the training data.
Such data shifts introduce additional epistemic uncertainty and can be characterized by the types of concepts they change and the degree to which concepts are changed, respectively. 
An example are the discrete 
data shifts that occur when model and uncertainty mechanism are trained on simulated inputs (as virtual open worlds allow \eg to generate arbitrarily many critical scenarios) and evaluated on real-world data. 
The (simulation-based) StreetHazards anomaly segmentation dataset (\cite{hendrycks2019scaling}) on the other side introduces high-level shifts as it modifies the semantics of traffic scenes by placing unknown unknowns (random objects like sheep and airplanes) into them.
For a given uncertainty use case, the data specifications of \eg the uncertainty ODD (see subsection \ref{subsec:modal_and_domain}) and the acceptance criteria (see subsection \ref{subsec:desiderata})
help to select relevant data shifts.
Low-level distortions like noising, blurring or changed colors and contrasts are of interest for most uncertainty estimators as they imitate flawed measurements and (likely) increase the aleatoric uncertainty of the input.

For ML models that process, for instance temporal, input streams, we test the (temporal) consistency of uncertainty estimates, \ie whether the change rates of uncertainty estimates reflect the input change rates.\footnote{If scenes are, for instance, only very slowly changing, one would expect the uncertainty estimates to decrease over time while sudden changes
in a scene would expectably lead to an increase in uncertainty.} 
This becomes especially important when (simple) algorithms for temporal smoothing like Kalman filters \citep{welch1995introduction} are employed that may limit uncertainty change rates.


\subsection{Subset and point-wise uncertainty tests}
\label{subsec:subset_tests}

An uncertainty estimator that is insufficient \wrt the above tests of ``global'' functioning is likely to be discarded.
Successfully passing those tests, however, is not a sufficient but only a necessary condition as they solely check aggregated uncertainty quality. Many uncertainty use cases, in contrast, require proper functioning in individual situations, especially in (potentially) critical ones. 

\begin{table}[t]
\centering
\caption{Four technical approaches to subset and point-wise uncertainty testing that are categorized based on data selection strategy (x-axis) and their (non-)semantic nature (y-axis), respectively. See subsection \ref{subsec:subset_tests} for details.}
\begin{tabular}[t]{>{\raggedright}p{0.25\linewidth}>{\raggedright}p{0.26\linewidth}>{\raggedright\arraybackslash}p{0.27\linewidth}}
\toprule
\addlinespace[0.7em]
\textit{subset and point-wise uncertainty testing} & \textbf{distributional} & \textbf{search-based / curated} \\
\addlinespace[0.5em]
\midrule
\addlinespace[0.7em]
\textbf{non-semantic} & output-uncertainty-based testing & testing with uncertainty-based search strategies \\
\addlinespace[0.5em]
\textbf{semantic} & testing along semantic dimensions & testing critical semantic input scenarios \\
\addlinespace[0.5em]
\bottomrule
\label{tab:subset_unc_test_matrix}
\end{tabular}
\vspace{-5mm}
\end{table}%

One way to study such critical inputs is revisiting the ``global'' tests, more concretely, their point-wise intermediate results (\eg normalized residuals or NLL values). 
Instead of averaging them (\eg by means of ECE or mean NLL), one may stay at a level of higher granularity and focus on individual quantiles or quantile ranges of these distributions. For details on such \textit{output-uncertainty-based testing}, see the paragraphs below. 

Datasets can not only be sliced based on numerical uncertainty scores but moreover based on high-level annotations, assuming that such information is available for each input data point. 
Accordingly derived \textit{semantic-dimension-based tests} (see below) reveal high-level strengths and weaknesses of an uncertainty mechanism and moreover allow to evaluate how well its domain of ``proper functioning'' covers the desired uncertainty ODD as specified in subsection \ref{subsec:modal_and_domain}. 

Going one step further, one may not only slice given datasets but curate (or even create) potentially critical inputs based on semantics. 
The sampling-based approach of determining the uncertainty ODD in subsection \ref{subsec:modal_and_domain} guides these data compilations that provide the basis for \textit{testing uncertainty properties in critical semantic scenarios} (see below).

Moreover, critical input scenarios may be generated by following ``hints'' on low local uncertainty quality. 
Such \textit{uncertainty-based search strategies} (see below) allow to explore the input space to uncover regions of insufficient ``self-assessment''. 
Given the sparsity of most datasets and the incomplete nature of semantic specifications, this test type poses a promising technical approach to reduce ``blind spots'' of subset and point-wise uncertainty testing.

\begin{figure}[t]
    \centering
    \includegraphics[trim=80 120 180 150, clip, width=0.95\textwidth]{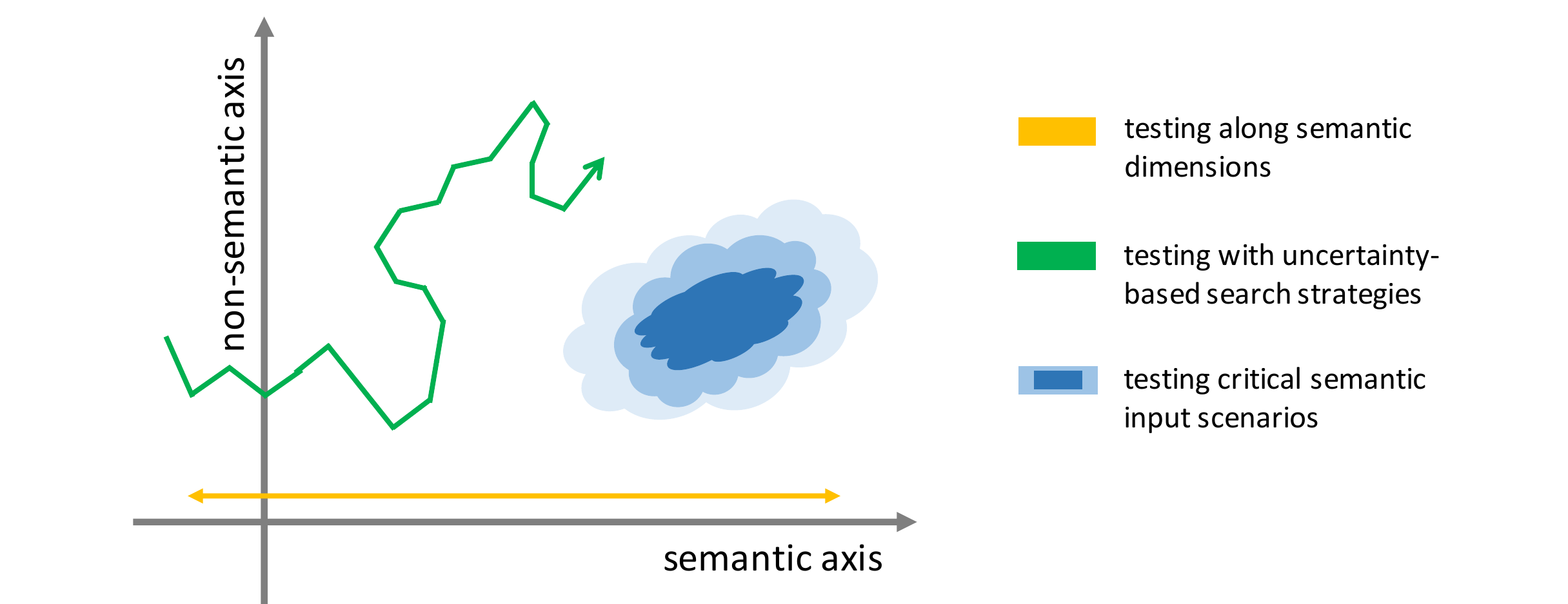}
    \caption{Symbolic illustration of three out of four data selection strategies on the testing hierarchy level of subset and point-wise tests. The data selection for output-uncertainty-based testing is trivial as it builds on the standard test dataset used for global testing (see above) and is thus not shown.
    The high-dimensional input space is sketched as a 2D plane with a semantic and a non-semantic axis. 
    Testing along a semantic dimension (orange arrow) requires (in most cases) a simulation environment that allows to systematically vary the corresponding input attribute while keeping all other input properties unchanged.
    However, in many real-world use cases, relevant regions of the input space (blue ``cloud'') are too complex to fully capture them this way. Instead, they can be represented \eg by expert-curated datasets.
    Search strategies finally do not require semantic information and may, for instance, be driven by gradients of local uncertainty quality (green trajectory). For details on all data selection strategies, see subsection \ref{subsec:subset_tests}.
    }
    \label{fig:input_selection_tests}
\end{figure}

We detail on these four testing techniques (see Tab.~\ref{tab:subset_unc_test_matrix}) in the following paragraphs. Fig.~\ref{fig:input_selection_tests} illustrates the differences between the sketched testing strategies by means of a symbolic 2D input space with a semantic and a non-semantic axis.

\paragraph{Output-uncertainty-based testing}

Slicing a dataset based on local output uncertainty quality requires a point-wise uncertainty measure, rendering set-based (calibration) metrics like ECE unsuited. 
More appropriate candidates are absolute values of uncertainty estimates and measures that put point-wise uncertainty estimates and (absolute) model errors into relation, like NLL values or normalized residuals.
The choice of the measure is guided by the uncertainty acceptance criteria (see section~\ref{sec:acceptance_criteria}).

Calculating the width of an uncertainty-output distribution (its variance or, for multi-dimensional output distributions, its covariance) provides insights on the meaningfulness of on-average uncertainty figures for individual input scenarios as these ``width'' scores measure the (average) deviation from mean behavior. 
Going beyond average deviations, we next focus on sub-datasets that cause strongly diverging uncertainty estimates and thus contribute to the tails of the uncertainty-output distribution.
For point-wise measures like NLL and normalized residuals, these tails correspond to uncertainty estimates that severely under- or overestimate the actual model errors. 
As discussed above, one may argue for the special practical relevance of under-estimated uncertainties, \ie of over-confident model predictions. 
These worst-case scenarios \wrt uncertainty quality are measured by quantile values (\eg $1\%$ or $99\%$) or measures that are sensitive to the ``depth'' of the distributional tails, such as conditioned mean values (\eg the ETL).
Since these measures allow to detect deviations from Gaussianity (see \eg \citet{sicking2020characteristics} for an analysis of these deviations), they may help to identify (overly) simplifying modeling assumptions and thus provide an additional safeguard. 

For ML models processing structured input, it may be insightful to attribute output uncertainty to input features. 
Input modifications like obfuscations allow to analyze the sensitivity of output uncertainties \wrt input regions (such as image patches, frequency bands or sub-graphs) and how susceptible an uncertainty estimator is \wrt local and global input features (\eg if or how strongly an object-level uncertainty changes when a far-off image patch is modified).

Relying on output uncertainty estimates, these tests do not require (high-level) meta-information. Being available however such semantic information may allow for an efficient and human-understandable description of the selected sub-datasets.
In contrast, the availability of high-level meta-information is a prerequisite for testing along semantic dimensions (see the next paragraph).


\paragraph{Testing along semantic dimensions}

Real-world ML systems likely encounter inputs at inference that are statistically novel compositions of known concepts (\eg of known object types)
and moreover structurally new concepts (unknown unknowns).
Analyzing how uncertainty quality changes in the presence of increased epistemic uncertainty due to varied scene semantics provides a high-level understanding of the estimator's extrapolation capabilities and its borders of functioning.
Concretely, one may study an ID-OOD transition along a semantic ``direction'',\footnote{We consider only modifications along one semantic ``axis'' at a time to avoid combinatorial explosions.} \eg whether object-level uncertainty estimates for VRUs are sufficiently calibrated as a function of their distance to the vehicle and how calibration quality may decrease for far away (and thus ``small'') VRUs. Another example of a semantic dimension in the field of AD are (labeled) body parts of VRUs that allow to analyze whether uncertainty quality diminishes for various kinds and degrees of VRU occlusion (another ID-OOD transition). 

Such tests allow to compare the actual abilities of an uncertainty estimator with its
ODD specification which \eg may ask for properly calibrated uncertainty estimates for VRUs that are within the breaking distance of a vehicle.
Technically, such analyses 
are based on
either richly annotated real-world data or simulated data. 
While the latter ones allow for a systematic variation along a semantic dimension as well as for ``zooming'' into critical sections (as arbitrary amounts of data can be generated), artificial data comes at the cost of a domain gap between, on the one hand, the accordingly conducted tests and, on the other hand, the actual usage of the ML model in the real world.

Moreover, these variations along only one semantic ``axis'' at a time
constrain the variability of the ``reachable'' inputs (one may think of a ``basis'' scene that is, one after the other, deflected in various directions).
While these semantic-axis tests provide a first understanding of high-level properties of an uncertainty estimator, further testing on more realistic input scenarios is required.


\paragraph{Testing critical semantic input scenarios}

Critical input scenarios are 
barely represented in standard datasets.
Moreover, they are often too complex to encounter them when varying a single semantic dimension of a basic input scenario using \eg a simulation engine (see previous ``paragraph'').
Both aspects, their rare occurrence and their complexity, render a decrease in model performance likely and thus large uncertainty estimates desirable.
Datasets of such critical input scenarios \citep{chan2016accidents,yao2019accidents,bao2020accidents} can either by curated by domain and safety experts (\eg \citet{jermakian2011primary}) or semi-automatically generated using \eg probabilistic scene grammars \citep{kar2019meta} that set up scenes based on pre-defined probability distributions.\footnote{The parameter distributions are either directly or indirectly (when learned from curated data) set by experts.}
The sampling-based approach to determining the ODD from subsection \ref{subsec:modal_and_domain} provides a notion of what constitutes a critical scene for an uncertainty use case and is thus a natural starting point for data curation or data modeling.
Also,
critical historical scenarios
can be included
if available (\eg data of historical human-induced crashes in the field of AD \citep{scanlon2021waymo}).

For ML models processing spatio-temporal data streams, one may moreover consider critical scenarios in the temporal domain such as abrupt changes of lighting conditions in the case of AD, \eg when driving in a tunnel or when the sun breaks through the clouds, and test whether uncertainty estimates change accordingly. 


\paragraph{Testing with uncertainty-based search strategies}

Available datasets typically capture only a small ``volume'' of the input space and thus reveal only a potentially small fraction of the actual weaknesses of an uncertainty estimator. 
To enlarge the data base, one could synthesize additional input points \eg by interpolating between or extrapolating from given test data points. 
More systematic however compared to such untargeted data acquisition strategies, are ``pro-active'' searches for input space regions that cause \eg large or ill-calibrated uncertainty estimates.

Various algorithmic search strategies to explore input data spaces exist (\eg \citet{pei2017deepxplore,odena2019tensorfuzz,klischat2019scenarios}). They (typically) follow ``hints'' of weak model performance \eg by means of gradient descent in a latent space.
Adversarial attacks fall into this category as well as coverage-based strategies (see related work in section~\ref{sec:related}). 
While most such approaches focus on model performance, we advocate for search strategies that are guided by uncertainty quality. They require point-wise uncertainty measures such as NLLs 
or normalized residuals and target larger regions of input space compared to performance testing due to the (in most cases) larger uncertainty ODD.
Depending on the type of uncertainty estimator, detected large uncertainty estimates can be attributed to either aleatoric or epistemic uncertainty.

Relying on numerical (point-wise) uncertainty scores, search-based tests (SBTs) do not require any meta-information, unlike testing along semantic dimensions and testing of critical input scenarios. 
However, non-semantic SBTs may still benefit from high-level annotations as the latter ones allow for (approximate) semantic descriptions of detected weaknesses (compare output-uncertainty\--based testing).


\subsection{Complementary uncertainty tests}
\label{subsec:complementary_tests}

The discussed testing approaches (see Tab.~\ref{tab:subset_unc_test_matrix}) are by no means definitive and further (especially application-specific) tests 
should be considered.
The complementary uncertainty tests are intended as an open residual set where as one example we present ``cross-analysis'' examinations.
We outline two variants of such tests: firstly, combinations of the uncertainty testing techniques discussed above,
\eg quantile range uncertainty analyses on a curated dataset of critical input scenarios. 
Secondly, ``cross-analysis'' tests in the sense that uncertainty estimates (and not the ``task outputs'' of the model) are analyzed \wrt other 
``dimensions'', such as interpretability or fairness.  
Considering \eg the fairness of uncertainty estimates, one may test whether uncertainty quality is comparably good across the values of sensitive input attributes.


\secsummary

The above presented testing hierarchy facilitates structured testing of uncertainty estimators where the sketched testing methodologies help to cover a broad range of relevant input space regions, paying attention to both their semantic and non-semantic aspects.
More concretely, we choose an appropriate test ``depth'' (in the test hierarchy) for each quantitative uncertainty acceptance criterion. For the respectively selected hierarchy levels relevant focus points of testing 
can be
determined \eg testing of low-level sensory distortions (on the hierarchy level of ``global tests'') and testing along semantic dimensions (on the hierarchy level of ``subset and point-wise tests''). 


\section{Instantiating, running and evaluating uncertainty test cases}
\label{sec:tests}

To enable statistical testing, the identified focus points of uncertainty testing (see previous section) need to be mapped onto concrete uncertainty test cases.
Best practices for formulating such test cases are outlined in subsection \ref{subsec:concrete_tests}.
Aspects of running the specified uncertainty tests are sketched in subsection \ref{subsec:operationalize_tests}. 
Given the (binary) results of the test cases, we provide guidelines on how to derive an overall evaluation of an
uncertainty estimator \wrt the uncertainty acceptance criteria in subsection \ref{subsec:evaluate_tests}.


\subsection{Instantiating uncertainty test cases}
\label{subsec:concrete_tests}

Uncertainty test cases can be understood as instantiations of high-level testing focuses (see section~\ref{sec:setup_unc_testing}). 
Similarly to the derivation of uncertainty acceptance criteria from qualitative requirements, these uncertainty tests are best formulated redundantly and collaboratively, \ie involving both domain and ML experts, to reduce ``blind spots''. Each testing focus is typically addressed by several test cases.
Guided by the uncertainty acceptance criteria, the data bases, threshold values\footnote{While the threshold value is in general given by the underlying acceptance criterion, it might be refined for some test cases, especially for subset-based ones. When investigating \eg worst-case scenarios of a given dataset, average values no longer hold.} and technical configurations (\eg the meta-parameters) for these test cases are determined.
Such specifications may vary between test cases, even between those that address the same acceptance criterion.
Uncertainty tests building on search strategies 
generate the test dataset during test execution and  thus do not require fully pre-specified datasets. Instead, their meta-parameters need to be set.

A prerequisite for a test to be conceptually meaningful, is its statistical significance that involves, among others, the following two aspects: 
first, whether a test dataset contains a broad set of relevant concepts (for the use and test case at hand) such that evidence-based conclusions can be drawn regarding the quality of the ML model at inference.
Second, whether the outcome of the test critically depends on low-level technical parameters such as random seeds for initialization. 
Insights in this regard may be gained by repeatedly executing the test in varying configurations and analyzing the sensitivities of the according test outcomes. 
Apart from statistical considerations, one may estimate the ``foot print'' of a test \wrt computation or storage, \ie the resources its preparation and execution consume. While the result in absolute terms may determine whether a test is affordable, relative results are helpful for test ordering and prioritization.

Comparing two test cases, each of their components (test dataset, uncertainty measure, threshold value, technical specification) might introduce overlaps or discrepancies between them. 
While certain overlaps are desirable (see above), we raise awareness for less obvious (implicit) dependencies between test cases:
These range from test data bases that have superset-subset relations or are strongly overlapping over measures of uncertainty quality that are correlated to threshold values that might be at odds.
Finally, one may analyze whether the ``union'' of all test cases sufficiently addresses the chosen focus points of testing or whether unaddressed ``blind spots'' remain.


\subsection{Running uncertainty test cases}
\label{subsec:operationalize_tests}

Since the uncertainty test cases are constructed to be (largely) independent of one another, they may be executed in an arbitrary order.
Following the structure of the test hierarchy, from general technical tests to specific complementary tests, may however be beneficial as potentially occurring severe weaknesses are detected at an early stage (in the worst case, ``fail fast'').
Having executed the uncertainty tests, their outcomes are uncertainty values that either exceed or fall below the threshold value of the respective test thus providing a binary test result (``passed'' or ``failed'').
In case of conflicting results for conceptually similar tests, it may be worthwhile to investigate whether the diverging outcomes are, at least to some extent, attributable to a specific difference in their test setups, \eg to varying sensitivities of uncertainty measures.
Such insights may result in better designed or additional test cases (compare respective backward arrow in Fig.~\ref{fig:process_diagram}).


\subsection{Evaluating uncertainty test cases}
\label{subsec:evaluate_tests}

Addressing uncertainty acceptance criteria by means of test focuses (organized in test levels) that are in turn mapped onto a set of test cases, introduces hierarchical tree-like structures.
Once the (final) set of tests is executed and all binary test results are available, these information at the ``leaves'' of the logical trees must be aggregated to obtain an overall evaluation of an uncertainty mechanism \wrt the acceptance criteria (the roots of the logical trees).
Aggregating the binary uncertainty test results and, in a second step, the uncertainty acceptance criteria requires a qualitative, highly application-specific argumentation that could be formulated using \eg a goal structuring notation \citep{kelly2004goal}. 
Depending on the uncertainty use case and the ``level'' of the acceptance criteria, such argumentations may be strict, \ie all acceptance criteria must be fulfilled (and maybe even all tests must be passed successfully) or more flexible, \ie a custom argumentation (employing \eg weightings of tests and a custom ``merging'' logic) allows for a positive overall evaluation despite a few non-fulfilled acceptance criteria (or at least despite a few failed tests).


\secsummary

Insights during testing and test result aggregation may trigger (iterative) adaptations of its scope and structure.
In the case that a ``converged'' testing strategy, \ie one leading to consistent test outcomes, results in an overall negative evaluation of an uncertainty estimator, uncertainty modeling may start all over again (compare respective backward arrows in Fig.~\ref{fig:process_diagram}). 
If, however, testing leads to an overall positive evaluation of an estimator \wrt the uncertainty acceptance criteria, the end point of the proposed development and testing framework is reached. 
In the case that high-level requirements change at a later date, e.g. due to external influences such as novel technical regulations,
it may be necessary to re-apply the uncertainty framework. In these situations, its modularity allows for the re-use of results from the previously conducted development and testing steps.


\section{Outlook}
\label{sec:discussion}

Uncertainty estimation can help to improve the safety of ML systems.
While various technical uncertainty mechanisms were put forward, less light is shed on how to systematically address application demands with them. We propose a framework that approaches uncertainty quantification from a practitioners' perspective, \ie starting from the various high-level requirements an uncertainty estimator is subject to. These range from the underlying use case that predetermines its desired properties and operational design domain to technical specifications that may limit its depth of integration into the ML model. For a more systematic analysis, each requirement is grouped into one of ten requirement categories. Formalizing the requirements as acceptance criteria with semantic data specification, uncertainty measure and threshold value yields the ``gold standard'' for the subsequent modeling and testing steps.
While a custom uncertainty modeling technique may be constructed to fulfill these uncertainty acceptance criteria, it is often worthwhile to use established uncertainty modeling techniques as ``building blocks''. We guide this construction of a \textit{tailored} estimator by matching the uncertainty requirement categories with central technical properties of uncertainty modeling (\eg modeled uncertainty types, generalization ability, attribution to uncertainty source).
The resulting requirements-informed uncertainty mechanism is systematically tested to uncover (potentially existing) structural weaknesses.
Technically, this is achieved by combinations of semantic and non-semantic testing approaches that are organized in a test hierarchy. For each acceptance criterion, a test ``depth'' (according to the hierarchy) and test focuses are determined. Instantiating these test focuses as concrete test cases allows to finally obtain (binary) test results. An overall evaluation of these results yields an answer whether the uncertainty estimator meets the uncertainty acceptance criteria, and thus, whether it is suitable for the given DL system.
These individual steps within the framework have well-defined hand-over points 
that allow for a strong degree of encapsulation.
This can be beneficial for larger projects where multiple mechanisms are being developed concurrently or when comparable testing approaches are needed for multiple products.

Furthermore, several steps of our conceptual framework can be operationalized and automatized in software frameworks. For instance, the selection of uncertainty models and metrics, as is done in existing frameworks like \textit{UQ360} \citep{ghosh2021uq360} and \textit{uncertainty-toolbox} \citep{chung2021uncertainty}.
Testing and artifact management may be made traceable and reproducible using ML lifecycle platforms like \textit{MLflow}\footnote{\url{https://github.com/mlflow/mlflow}.}.
Desirable enhancements of such tools might moreover contain components such as requirement templates,
building blocks for qualitative (safety) argumentations and backlogs of standard tests.
Particularly, the (semi-)automation and optimization of quantitative testing bears potential. 
Promising technical approaches in this regard
range from ML-based strategies to speed-up a concrete test case (\eg by predicting, right after starting the test, whether its outcome will be negative),
over the (\eg Bayesian) generation of the best follow-up test configuration for a pre-defined test type to the development of heuristics on how to effectively react on failing test cases. 
Such software frameworks may 
moreover enable effective test-driven design and development of uncertainty estimators. Gained insights and uncovered weaknesses could even fuel research on novel uncertainty modeling techniques.

While our framework structures the testing of a single model, industry applications often have a multi-step development approach. It might be worthwhile to investigate to which extend early, and therefore easier to obtain, test results are indicative for success or failure of later stages of development. The proposed encapsulation of testing from development might help in the necessary transfer of results. Achieving a strong correlation, for instance with respect to qualitative uncertainty properties, may help to streamline the development process or to identify critical conceptual building blocks of the DL system.

Technical requirements aside, various official regulations for learned systems are about to emerge, especially in the EU \citep{eu2021regulation}, where, likely from 2025 on, newly created and substantially changed DL applications will be regulated by an ``AI Act'' if the respective application type is considered to be critical. Our framework may be of help for associated product and process examinations, especially those concerning model reliability and the handling of unknown inputs.
It might also serve as a structural aid for the technical documentations required by the prospective \mbox{EU regulations.}

While uncertainty estimates help to fulfill requirements for trustworthy ML, there are multiple other ways to either increase reliability, \eg robustified training \citep{madry2017towards}, or other aspects of trustworthiness, for instance fairness or interpretability. The proposed
framework, despite its specific focus on uncertainty quantification, remains applicable in these cases.
The main difficulties it addresses, namely the black-box nature of the neural models and their high-dimensional input spaces, are common to most safe-ML techniques and approaches.
The proposed hierarchical test strategy, for instance, is transferable as it addresses the high-dimensional input spaces by specific combinations of (semantic and non-semantic) testing methodologies and test data selection routines.
Moreover, the ways how underlying use case and technical specifications predetermine the conception and implementation of the safe-ML techniques carry over to other ``dimensions'' of safe ML.  
At a level of technical properties, however,
some requirement categories like ``calibration'' or ``attribution to type of uncertainty'' as well as mechanisms and measures are uncertainty-specific and need to be replaced by appropriate (statistical) concepts for the respective ``dimension'' of safe ML (\eg by ``interpretable-by-design'' for qualitative interpretability requirements or ``group and individual fairness'' as a quantitative demand for fair ML models).

Such procedures for developing and testing safe-ML tools may be complemented, especially in larger organizations, by a role-based access and rights management that builds on the encapsulation of our approach and enhances the interfaces between the various steps and the respectively involved teams (\eg development and product). 
Distributing responsibilities in that way seems especially advantageous for the development and testing of ML components:
development teams, for instance, may not have access to the concrete test cases to avoid model optimization that ``overfits'' to these cases. Alternatively, a part of the test cases may be hidden from the developers while others are still accessible to them. This may be compared to the
public and private datasets of ML challenges.
Taking inspiration from IT security \citep{roy2010cyber,zhang2019attacker}, testing teams could moreover act as ``attackers'' that continuously challenge the uncertainty estimators developed by the modeling team, a competitive serious game that may result in more robust ML systems.  

In many real-world applications, the deployed DL systems are not static but further improve during operations,
either due to continuous learning or, and more likely for most safety-critical applications, after scheduled updates that are reviewed before going into production. 
For such an updated model the question arises whether testing and validation needs to be done all over again or whether existing test results from previous versions of the model can be re-used.
In particular, an interesting aspect could be to figure out whether mild deviations in high-level tests of the hierarchy can yield insights on the ``transferability'' of previous results on the deeper parts of the hierarchy.
Should this hold the testing hierarchy may contribute to more efficient model validation in the long run.

By proposing the presented framework, we contribute a holistic, application-driven perspective on uncertainty estimation in deep learning models.
We hope that our guidelines assist in a more appropriate choice or development of uncertainty estimators, allowing for a use of this promising modeling technique that is tailored to match application demands.
The multi-staged testing procedure may not only help to satisfy upcoming ML regulations but may increase the overall reliability, and thus the quality, of the DL systems.


\section*{Acknowledgments}
The research of J.\ Sicking and T.\ Wirtz was funded by the Fraunhofer Center for Machine Learning within the Fraunhofer Cluster for Cognitive Internet Technologies. The work of M.\ Akila was funded by the German Federal Ministry of Education and Research, ML2R - no. 01S18038B. The authors thank M. Poretschkin for fruitful discussions on the standardization and regulation of ML systems.


\bibliography{references}

\begin{thebibliography}{117}
\providecommand{\natexlab}[1]{#1}
\providecommand{\url}[1]{\texttt{#1}}
\expandafter\ifx\csname urlstyle\endcsname\relax
  \providecommand{\doi}[1]{doi: #1}\else
  \providecommand{\doi}{doi: \begingroup \urlstyle{rm}\Url}\fi

\bibitem[Abdar et~al.(2021)Abdar, Pourpanah, Hussain, Rezazadegan, Liu,
  Ghavamzadeh, Fieguth, Cao, Khosravi, Acharya, Makarenkov, and
  Nahavandi]{abdar2021review}
Moloud Abdar, Farhad Pourpanah, Sadiq Hussain, Dana Rezazadegan, Li~Liu,
  Mohammad Ghavamzadeh, Paul~W. Fieguth, Xiaochun Cao, Abbas Khosravi,
  U.~Rajendra Acharya, Vladimir Makarenkov, and Saeid Nahavandi.
\newblock A review of uncertainty quantification in deep learning: Techniques,
  applications and challenges.
\newblock \emph{Inf. Fusion}, 76:\penalty0 243--297, 2021.
\newblock \doi{10.1016/j.inffus.2021.05.008}.
\newblock URL \url{https://doi.org/10.1016/j.inffus.2021.05.008}.

\bibitem[Adel et~al.(2019)Adel, Valera, Ghahramani, and
  Weller]{adel2019fairness}
Tameem Adel, Isabel Valera, Zoubin Ghahramani, and Adrian Weller.
\newblock One-network adversarial fairness.
\newblock In \emph{The Thirty-Third {AAAI} Conference on Artificial
  Intelligence, {AAAI} 2019, The Thirty-First Innovative Applications of
  Artificial Intelligence Conference, {IAAI} 2019, The Ninth {AAAI} Symposium
  on Educational Advances in Artificial Intelligence, {EAAI} 2019, Honolulu,
  Hawaii, USA, January 27 - February 1, 2019}, pp.\  2412--2420. {AAAI} Press,
  2019.
\newblock \doi{10.1609/aaai.v33i01.33012412}.
\newblock URL \url{https://doi.org/10.1609/aaai.v33i01.33012412}.

\bibitem[Amini et~al.(2020)Amini, Schwarting, Soleimany, and
  Rus]{amini2020evidential}
Alexander Amini, Wilko Schwarting, Ava Soleimany, and Daniela Rus.
\newblock Deep evidential regression.
\newblock In H.~Larochelle, M.~Ranzato, R.~Hadsell, M.~F. Balcan, and H.~Lin
  (eds.), \emph{Advances in Neural Information Processing Systems}, volume~33,
  pp.\  14927--14937. Curran Associates, Inc., 2020.
\newblock URL
  \url{https://proceedings.neurips.cc/paper/2020/file/aab085461de182608ee9f607f3f7d18f-Paper.pdf}.

\bibitem[Amodei et~al.(2016)Amodei, Olah, Steinhardt, Christiano, Schulman, and
  Man{\'{e}}]{amodei2016concrete}
Dario Amodei, Chris Olah, Jacob Steinhardt, Paul~F. Christiano, John Schulman,
  and Dan Man{\'{e}}.
\newblock Concrete problems in {AI} safety.
\newblock \emph{CoRR}, abs/1606.06565, 2016.
\newblock URL \url{http://arxiv.org/abs/1606.06565}.

\bibitem[Aravantinos \& Schlicht(2020)Aravantinos and
  Schlicht]{aravantinos2020uncertainty}
Vincent Aravantinos and Peter Schlicht.
\newblock Making the relationship between uncertainty estimation and safety
  less uncertain.
\newblock In \emph{2020 Design, Automation {\&} Test in Europe Conference {\&}
  Exhibition, {DATE} 2020, Grenoble, France, March 9-13, 2020}, pp.\
  1139--1144. {IEEE}, 2020.
\newblock \doi{10.23919/DATE48585.2020.9116541}.
\newblock URL \url{https://doi.org/10.23919/DATE48585.2020.9116541}.

\bibitem[Arnold et~al.(2019)Arnold, Bellamy, Hind, Houde, Mehta, Mojsilovic,
  Nair, Ramamurthy, Olteanu, Piorkowski, Reimer, Richards, Tsay, and
  Varshney]{arnold2019factsheets}
Matthew Arnold, Rachel K.~E. Bellamy, Michael Hind, Stephanie Houde, Sameep
  Mehta, Aleksandra Mojsilovic, Ravi Nair, Karthikeyan~Natesan Ramamurthy,
  Alexandra Olteanu, David Piorkowski, Darrell Reimer, John~T. Richards, Jason
  Tsay, and Kush~R. Varshney.
\newblock Factsheets: Increasing trust in {AI} services through supplier's
  declarations of conformity.
\newblock \emph{{IBM} J. Res. Dev.}, 63\penalty0 (4/5):\penalty0 6:1--6:13,
  2019.
\newblock \doi{10.1147/JRD.2019.2942288}.
\newblock URL \url{https://doi.org/10.1147/JRD.2019.2942288}.

\bibitem[Bao et~al.(2020)Bao, Yu, and Kong]{bao2020accidents}
Wentao Bao, Qi~Yu, and Yu~Kong.
\newblock Uncertainty-based traffic accident anticipation with spatio-temporal
  relational learning.
\newblock In Chang~Wen Chen, Rita Cucchiara, Xian{-}Sheng Hua, Guo{-}Jun Qi,
  Elisa Ricci, Zhengyou Zhang, and Roger Zimmermann (eds.), \emph{{MM} '20: The
  28th {ACM} International Conference on Multimedia, Virtual Event / Seattle,
  WA, USA, October 12-16, 2020}, pp.\  2682--2690. {ACM}, 2020.
\newblock \doi{10.1145/3394171.3413827}.
\newblock URL \url{https://doi.org/10.1145/3394171.3413827}.

\bibitem[Braiek \& Khomh(2020)Braiek and Khomh]{braiek2020testing}
Houssem~Ben Braiek and Foutse Khomh.
\newblock On testing machine learning programs.
\newblock \emph{J. Syst. Softw.}, 164:\penalty0 110542, 2020.
\newblock \doi{10.1016/j.jss.2020.110542}.
\newblock URL \url{https://doi.org/10.1016/j.jss.2020.110542}.

\bibitem[Brendel \& Bethge(2019)Brendel and Bethge]{brendel2019approximating}
Wieland Brendel and Matthias Bethge.
\newblock Approximating {CNNs} with bag-of-local-features models works
  surprisingly well on {ImageNet}.
\newblock In \emph{7th International Conference on Learning Representations,
  {ICLR} 2019, New Orleans, LA, USA, May 6-9, 2019}. OpenReview.net, 2019.
\newblock URL \url{https://openreview.net/forum?id=SkfMWhAqYQ}.

\bibitem[Chan et~al.(2016)Chan, Chen, Xiang, and Sun]{chan2016accidents}
Fu{-}Hsiang Chan, Yu{-}Ting Chen, Yu~Xiang, and Min Sun.
\newblock Anticipating accidents in dashcam videos.
\newblock In Shang{-}Hong Lai, Vincent Lepetit, Ko~Nishino, and Yoichi Sato
  (eds.), \emph{Computer Vision - {ACCV} 2016 - 13th Asian Conference on
  Computer Vision, Taipei, Taiwan, November 20-24, 2016, Revised Selected
  Papers, Part {IV}}, volume 10114 of \emph{Lecture Notes in Computer Science},
  pp.\  136--153. Springer, 2016.
\newblock \doi{10.1007/978-3-319-54190-7\_9}.
\newblock URL \url{https://doi.org/10.1007/978-3-319-54190-7\_9}.

\bibitem[Chan et~al.(2019)Chan, Rottmann, H{\"{u}}ger, Schlicht, and
  Gottschalk]{chan2019decisionrules}
Robin Chan, Matthias Rottmann, Fabian H{\"{u}}ger, Peter Schlicht, and Hanno
  Gottschalk.
\newblock Application of decision rules for handling class imbalance in
  semantic segmentation.
\newblock \emph{CoRR}, abs/1901.08394, 2019.
\newblock URL \url{http://arxiv.org/abs/1901.08394}.

\bibitem[Chrystal et~al.(2003)Chrystal, Mizen, and Mizen]{chrystal2003goodhart}
K~Alec Chrystal, Paul~D Mizen, and PD~Mizen.
\newblock Goodhart’s law: its origins, meaning and implications for monetary
  policy.
\newblock \emph{Central banking, monetary theory and practice: Essays in honour
  of Charles Goodhart}, 1:\penalty0 221--243, 2003.

\bibitem[Chung et~al.(2021)Chung, Char, Guo, Schneider, and
  Neiswanger]{chung2021uncertainty}
Youngseog Chung, Ian Char, Han Guo, Jeff Schneider, and Willie Neiswanger.
\newblock Uncertainty toolbox: an open-source library for assessing,
  visualizing, and improving uncertainty quantification.
\newblock \emph{CoRR}, abs/2109.10254, 2021.
\newblock URL \url{https://arxiv.org/abs/2109.10254}.

\bibitem[DeGroot \& Fienberg(1983)DeGroot and Fienberg]{degroot1983comparison}
Morris~H DeGroot and Stephen~E Fienberg.
\newblock The comparison and evaluation of forecasters.
\newblock \emph{Journal of the Royal Statistical Society: Series D (The
  Statistician)}, 32\penalty0 (1-2):\penalty0 12--22, 1983.

\bibitem[Durasov et~al.(2021)Durasov, Bagautdinov, Baqu{\'{e}}, and
  Fua]{durasov2021masksembles}
Nikita Durasov, Timur~M. Bagautdinov, Pierre Baqu{\'{e}}, and Pascal Fua.
\newblock Masksembles for uncertainty estimation.
\newblock In \emph{{IEEE} Conference on Computer Vision and Pattern
  Recognition, {CVPR} 2021, virtual, June 19-25, 2021}, pp.\  13539--13548.
  Computer Vision Foundation / {IEEE}, 2021.
\newblock URL
  \url{https://openaccess.thecvf.com/content/CVPR2021/html/Durasov\_Masksembles\_for\_Uncertainty\_Estimation\_CVPR\_2021\_paper.html}.

\bibitem[Dwyer et~al.(1999)Dwyer, Avrunin, and
  Corbett]{dwyer1999finitestateverification}
Matthew~B. Dwyer, George~S. Avrunin, and James~C. Corbett.
\newblock Patterns in property specifications for finite-state verification.
\newblock In Barry~W. Boehm, David Garlan, and Jeff Kramer (eds.),
  \emph{Proceedings of the 1999 International Conference on Software
  Engineering, ICSE' 99, Los Angeles, CA, USA, May 16-22, 1999}, pp.\
  411--420. {ACM}, 1999.
\newblock \doi{10.1145/302405.302672}.
\newblock URL \url{https://doi.org/10.1145/302405.302672}.

\bibitem[Esteve et~al.(2012)Esteve, Katoen, Nguyen, Postma, and
  Yushtein]{esteve2012satellite}
Marie{-}Aude Esteve, Joost{-}Pieter Katoen, Viet~Yen Nguyen, Bart Postma, and
  Yuri Yushtein.
\newblock Formal correctness, safety, dependability, and performance analysis
  of a satellite.
\newblock In Martin Glinz, Gail~C. Murphy, and Mauro Pezz{\`{e}} (eds.),
  \emph{34th International Conference on Software Engineering, June 2-9, 2012,
  Zurich, Switzerland}, pp.\  1022--1031. {IEEE} Computer Society, 2012.
\newblock \doi{10.1109/ICSE.2012.6227118}.
\newblock URL \url{https://doi.org/10.1109/ICSE.2012.6227118}.

\bibitem[{European Commission}(2021)]{eu2021regulation}
{European Commission}.
\newblock {Proposal for a regulation laying down harmonised rules on artificial
  intelligence}, 2021.
\newblock URL
  \url{https://digital-strategy.ec.europa.eu/en/library/proposal-regulation-laying-down-harmonised-rules-artificial-intelligence}.
\newblock (accessed 26.05.2021).

\bibitem[{Executive Office of the President}(2020)]{whitehouse2020guidance}
{Executive Office of the President}.
\newblock Guidance for regulation of artificial intelligence applications,
  2020.
\newblock URL
  \url{https://www.whitehouse.gov/wp-content/uploads/2020/11/M-21-06.pdf}.

\bibitem[Gal \& Ghahramani(2016)Gal and Ghahramani]{gal2016dropout}
Yarin Gal and Zoubin Ghahramani.
\newblock Dropout as a {B}ayesian approximation: Representing model uncertainty
  in deep learning.
\newblock In Maria~Florina Balcan and Kilian~Q. Weinberger (eds.),
  \emph{Proceedings of The 33rd International Conference on Machine Learning},
  volume~48 of \emph{Proceedings of Machine Learning Research}, pp.\
  1050--1059, New York, New York, USA, 20--22 Jun 2016. PMLR.
\newblock URL \url{http://proceedings.mlr.press/v48/gal16.html}.

\bibitem[Gannamaneni et~al.(2021)Gannamaneni, Houben, and
  Akila]{gannamaneni2021concepttesting}
Sujan~Sai Gannamaneni, Sebastian Houben, and Maram Akila.
\newblock Semantic concept testing in autonomous driving by extraction of
  object-level annotations from {CARLA}.
\newblock In \emph{{IEEE/CVF} International Conference on Computer Vision
  Workshops, {ICCVW} 2021, Montreal, BC, Canada, October 11-17, 2021}, pp.\
  1006--1014. {IEEE}, 2021.
\newblock \doi{10.1109/ICCVW54120.2021.00117}.
\newblock URL \url{https://doi.org/10.1109/ICCVW54120.2021.00117}.

\bibitem[Gawlikowski et~al.(2021)Gawlikowski, Tassi, Ali, Lee, Humt, Feng,
  Kruspe, Triebel, Jung, Roscher, Shahzad, Yang, Bamler, and
  Zhu]{gawlikowski2021survey}
Jakob Gawlikowski, Cedrique Rovile~Njieutcheu Tassi, Mohsin Ali, Jongseok Lee,
  Matthias Humt, Jianxiang Feng, Anna~M. Kruspe, Rudolph Triebel, Peter Jung,
  Ribana Roscher, Muhammad Shahzad, Wen Yang, Richard Bamler, and Xiao~Xiang
  Zhu.
\newblock A survey of uncertainty in deep neural networks.
\newblock \emph{CoRR}, abs/2107.03342, 2021.
\newblock URL \url{https://arxiv.org/abs/2107.03342}.

\bibitem[Ghosh et~al.(2021)Ghosh, Liao, Ramamurthy, Navr{\'{a}}til, Sattigeri,
  Varshney, and Zhang]{ghosh2021uq360}
Soumya Ghosh, Q.~Vera Liao, Karthikeyan~Natesan Ramamurthy, Jir{\'{\i}}
  Navr{\'{a}}til, Prasanna Sattigeri, Kush~R. Varshney, and Yunfeng Zhang.
\newblock Uncertainty quantification 360: {A} holistic toolkit for quantifying
  and communicating the uncertainty of {AI}.
\newblock \emph{CoRR}, abs/2106.01410, 2021.
\newblock URL \url{https://arxiv.org/abs/2106.01410}.

\bibitem[Giannetti(2017)]{giannetti2017uncertaintyindustry}
Cinzia Giannetti.
\newblock A framework for improving process robustness with quantification of
  uncertainties in industry 4.0.
\newblock In Piotr Jedrzejowicz, T{\"{u}}lay Yildirim, and Ireneusz Czarnowski
  (eds.), \emph{{IEEE} International Conference on INnovations in Intelligent
  SysTems and Applications, {INISTA} 2017, Gdynia, Poland, July 3-5, 2017},
  pp.\  189--194. {IEEE}, 2017.
\newblock \doi{10.1109/INISTA.2017.8001155}.
\newblock URL \url{https://doi.org/10.1109/INISTA.2017.8001155}.

\bibitem[Gneiting \& Raftery(2007)Gneiting and Raftery]{gneiting2007strictly}
Tilmann Gneiting and Adrian~E Raftery.
\newblock Strictly proper scoring rules, prediction, and estimation.
\newblock \emph{Journal of the American Statistical Association}, 102\penalty0
  (477):\penalty0 359--378, 2007.
\newblock \doi{10.1198/016214506000001437}.
\newblock URL \url{https://doi.org/10.1198/016214506000001437}.

\bibitem[Goodfellow et~al.(2015)Goodfellow, Shlens, and
  Szegedy]{goodfellow2014adversarialexamples}
Ian~J. Goodfellow, Jonathon Shlens, and Christian Szegedy.
\newblock Explaining and harnessing adversarial examples.
\newblock In Yoshua Bengio and Yann LeCun (eds.), \emph{3rd International
  Conference on Learning Representations, {ICLR} 2015, San Diego, CA, USA, May
  7-9, 2015, Conference Track Proceedings}, 2015.
\newblock URL \url{http://arxiv.org/abs/1412.6572}.

\bibitem[Grunske(2008)]{grunske2008specificationpatterns}
Lars Grunske.
\newblock Specification patterns for probabilistic quality properties.
\newblock In Wilhelm Sch{\"{a}}fer, Matthew~B. Dwyer, and Volker Gruhn (eds.),
  \emph{30th International Conference on Software Engineering {(ICSE} 2008),
  Leipzig, Germany, May 10-18, 2008}, pp.\  31--40. {ACM}, 2008.
\newblock \doi{10.1145/1368088.1368094}.
\newblock URL \url{https://doi.org/10.1145/1368088.1368094}.

\bibitem[Guo et~al.(2017)Guo, Pleiss, Sun, and Weinberger]{guo2017calibration}
Chuan Guo, Geoff Pleiss, Yu~Sun, and Kilian~Q. Weinberger.
\newblock On calibration of modern neural networks.
\newblock In Doina Precup and Yee~Whye Teh (eds.), \emph{Proceedings of the
  34th International Conference on Machine Learning, {ICML} 2017, Sydney, NSW,
  Australia, 6-11 August 2017}, volume~70 of \emph{Proceedings of Machine
  Learning Research}, pp.\  1321--1330. {PMLR}, 2017.
\newblock URL \url{http://proceedings.mlr.press/v70/guo17a.html}.

\bibitem[Hadfield{-}Menell et~al.(2017)Hadfield{-}Menell, Milli, Abbeel,
  Russell, and Dragan]{hadfield2017inversereward}
Dylan Hadfield{-}Menell, Smitha Milli, Pieter Abbeel, Stuart~J. Russell, and
  Anca~D. Dragan.
\newblock Inverse reward design.
\newblock In Isabelle Guyon, Ulrike von Luxburg, Samy Bengio, Hanna~M. Wallach,
  Rob Fergus, S.~V.~N. Vishwanathan, and Roman Garnett (eds.), \emph{Advances
  in Neural Information Processing Systems 30: Annual Conference on Neural
  Information Processing Systems 2017, December 4-9, 2017, Long Beach, CA,
  {USA}}, pp.\  6765--6774, 2017.
\newblock URL
  \url{https://proceedings.neurips.cc/paper/2017/hash/32fdab6559cdfa4f167f8c31b9199643-Abstract.html}.

\bibitem[Hall et~al.(2020)Hall, Dayoub, Skinner, Zhang, Miller, Corke,
  Carneiro, Angelova, and S{\"{u}}nderhauf]{hall2020probabilistic}
David Hall, Feras Dayoub, John Skinner, Haoyang Zhang, Dimity Miller, Peter
  Corke, Gustavo Carneiro, Anelia Angelova, and Niko S{\"{u}}nderhauf.
\newblock Probabilistic object detection: Definition and evaluation.
\newblock In \emph{{IEEE} Winter Conference on Applications of Computer Vision,
  {WACV} 2020, Snowmass Village, CO, USA, March 1-5, 2020}, pp.\  1020--1029.
  {IEEE}, 2020.
\newblock \doi{10.1109/WACV45572.2020.9093599}.
\newblock URL \url{https://doi.org/10.1109/WACV45572.2020.9093599}.

\bibitem[Hartman(2006)]{hartman2006testing}
Alan Hartman.
\newblock Software and hardware testing using combinatorial covering suites.
\newblock \emph{Journal of Graph Theory - JGT}, 34:\penalty0 237--266, 03 2006.
\newblock \doi{10.1007/0-387-25036-0_10}.

\bibitem[He et~al.(2019)He, Zhu, Wang, Savvides, and Zhang]{he2019bounding}
Yihui He, Chenchen Zhu, Jianren Wang, Marios Savvides, and Xiangyu Zhang.
\newblock Bounding box regression with uncertainty for accurate object
  detection.
\newblock In \emph{{IEEE} Conference on Computer Vision and Pattern
  Recognition, {CVPR} 2019, Long Beach, CA, USA, June 16-20, 2019}, pp.\
  2888--2897. Computer Vision Foundation / {IEEE}, 2019.
\newblock \doi{10.1109/CVPR.2019.00300}.
\newblock URL
  \url{http://openaccess.thecvf.com/content_CVPR_2019/html/He_Bounding_Box_Regression_With_Uncertainty_for_Accurate_Object_Detection_CVPR_2019_paper.html}.

\bibitem[Henaff et~al.(2019)Henaff, Canziani, and LeCun]{henaff2019model}
Mikael Henaff, Alfredo Canziani, and Yann LeCun.
\newblock Model-predictive policy learning with uncertainty regularization for
  driving in dense traffic.
\newblock In \emph{7th International Conference on Learning Representations,
  {ICLR} 2019, New Orleans, LA, USA, May 6-9, 2019}. OpenReview.net, 2019.
\newblock URL \url{https://openreview.net/forum?id=HygQBn0cYm}.

\bibitem[Hendrycks et~al.(2019)Hendrycks, Basart, Mazeika, Mostajabi,
  Steinhardt, and Song]{hendrycks2019scaling}
Dan Hendrycks, Steven Basart, Mantas Mazeika, Mohammadreza Mostajabi, Jacob
  Steinhardt, and Dawn Song.
\newblock Scaling out-of-distribution detection for real-world settings.
\newblock \emph{arXiv preprint arXiv:1911.11132}, 2019.

\bibitem[Holstein et~al.(2019)Holstein, Vaughan, III, Dud{\'{\i}}k, and
  Wallach]{holstein2019improving}
Kenneth Holstein, Jennifer~Wortman Vaughan, Hal~Daum{\'{e}} III, Miroslav
  Dud{\'{\i}}k, and Hanna~M. Wallach.
\newblock Improving fairness in machine learning systems: What do industry
  practitioners need?
\newblock In Stephen~A. Brewster, Geraldine Fitzpatrick, Anna~L. Cox, and
  Vassilis Kostakos (eds.), \emph{Proceedings of the 2019 {CHI} Conference on
  Human Factors in Computing Systems, {CHI} 2019, Glasgow, Scotland, UK, May
  04-09, 2019}, pp.\  600. {ACM}, 2019.
\newblock \doi{10.1145/3290605.3300830}.
\newblock URL \url{https://doi.org/10.1145/3290605.3300830}.

\bibitem[Houben et~al.(2021)Houben, Abrecht, Akila, B{\"a}r, Brockherde,
  Feifel, Fingscheidt, Gannamaneni, Ghobadi, Hammam, et~al.]{houben2021inspect}
Sebastian Houben, Stephanie Abrecht, Maram Akila, Andreas B{\"a}r, Felix
  Brockherde, Patrick Feifel, Tim Fingscheidt, Sujan~Sai Gannamaneni,
  Seyed~Eghbal Ghobadi, Ahmed Hammam, et~al.
\newblock Inspect, understand, overcome: A survey of practical methods for {AI}
  safety.
\newblock \emph{arXiv preprint arXiv:2104.14235}, 2021.

\bibitem[H{\"{u}}llermeier \& Waegeman(2021)H{\"{u}}llermeier and
  Waegeman]{huellermeier2021taxonomy}
Eyke H{\"{u}}llermeier and Willem Waegeman.
\newblock Aleatoric and epistemic uncertainty in machine learning: an
  introduction to concepts and methods.
\newblock \emph{Mach. Learn.}, 110\penalty0 (3):\penalty0 457--506, 2021.
\newblock \doi{10.1007/s10994-021-05946-3}.
\newblock URL \url{https://doi.org/10.1007/s10994-021-05946-3}.

\bibitem[Ilg et~al.(2018)Ilg, {\c{C}}i{\c{c}}ek, Galesso, Klein, Makansi,
  Hutter, and Brox]{ilg2018uncertainty}
Eddy Ilg, {\"{O}}zg{\"{u}}n {\c{C}}i{\c{c}}ek, Silvio Galesso, Aaron Klein,
  Osama Makansi, Frank Hutter, and Thomas Brox.
\newblock Uncertainty estimates and multi-hypotheses networks for optical flow.
\newblock In Vittorio Ferrari, Martial Hebert, Cristian Sminchisescu, and Yair
  Weiss (eds.), \emph{Computer Vision - {ECCV} 2018 - 15th European Conference,
  Munich, Germany, September 8-14, 2018, Proceedings, Part {VII}}, volume 11211
  of \emph{Lecture Notes in Computer Science}, pp.\  677--693. Springer, 2018.
\newblock \doi{10.1007/978-3-030-01234-2\_40}.
\newblock URL \url{https://doi.org/10.1007/978-3-030-01234-2\_40}.

\bibitem[Ishikawa \& Yoshioka(2019)Ishikawa and
  Yoshioka]{ishikawa2019engineers}
Fuyuki Ishikawa and Nobukazu Yoshioka.
\newblock How do engineers perceive difficulties in engineering of
  machine-learning systems?: questionnaire survey.
\newblock In Marcus Ciolkowski, Dusica Marijan, Matthias Galster, Weiyi Shang,
  Andreas Jedlitschka, Rakesh Shukla, and Kanchana Padmanabhan (eds.),
  \emph{Proceedings of the Joint 7th International Workshop on Conducting
  Empirical Studies in Industry and 6th International Workshop on Software
  Engineering Research and Industrial Practice, CESSER-IP@ICSE 2019, Montreal,
  QC, Canada, May 27, 2019}, pp.\  2--9. {IEEE} / {ACM}, 2019.
\newblock \doi{10.1109/CESSER-IP.2019.00009}.
\newblock URL \url{https://dl.acm.org/citation.cfm?id=3338708}.

\bibitem[{ISO TC 22/SC 32}(2021)]{iso2021safe-ai-road-vehicle}
{ISO TC 22/SC 32}.
\newblock {ISO/AWI PAS 8800}: Road vehicles — safety and artificial
  intelligence, 2021.
\newblock URL \url{https://www.iso.org/standard/83303.html}.

\bibitem[{ISO/IEC JTC 1/SC 42}(2017)]{iso2017ai}
{ISO/IEC JTC 1/SC 42}.
\newblock Standardization in the area of artificial intelligence, 2017.
\newblock URL \url{https://www.iso.org/committee/6794475.html}.

\bibitem[{ISO/IEC JTC 1/SC 42}(2020)]{iso2020trustworthyai}
{ISO/IEC JTC 1/SC 42}.
\newblock {ISO/IEC TR 24028:2020}: Information technology - artificial
  intelligence - overview of trustworthiness in artificial intelligence, 2020.
\newblock URL \url{https://www.iso.org/standard/77608.html}.

\bibitem[{ISO/IEC JTC 1/SC 42}(2021{\natexlab{a}})]{iso2021assessrobustness}
{ISO/IEC JTC 1/SC 42}.
\newblock {ISO/IEC TR 24029-1:2021}: Artificial intelligence - assessment of
  the robustness of neural networks - part 1: Overview, 2021{\natexlab{a}}.
\newblock URL \url{https://www.iso.org/standard/77609.html}.

\bibitem[{ISO/IEC JTC 1/SC 42}(2021{\natexlab{b}})]{iso2021managementai}
{ISO/IEC JTC 1/SC 42}.
\newblock {ISO/IEC CD 42001}: Information technology - artificial intelligence
  - management system, 2021{\natexlab{b}}.
\newblock URL \url{https://www.iso.org/standard/81230.html}.

\bibitem[{ISO/IEC JTC 1/SC 7}(2014)]{iso2014softwarequality}
{ISO/IEC JTC 1/SC 7}.
\newblock {ISO/IEC 25000:2014}: Systems and software engineering - systems and
  software quality requirements and evaluation (square) - guide to square,
  2014.
\newblock URL \url{https://www.iso.org/standard/64764.html}.

\bibitem[{ISO/IEC JTC 1/SC 7}(2019)]{iso2019assurance}
{ISO/IEC JTC 1/SC 7}.
\newblock {ISO/IEC/IEEE 15026-1:2019}: Systems and software engineering -
  systems and software assurance - part 1: Concepts and vocabulary, 2019.
\newblock URL \url{https://www.iso.org/standard/73567.html}.

\bibitem[{ISO/TC 22/SC 32}(2018)]{iso2018functionalsafety}
{ISO/TC 22/SC 32}.
\newblock {ISO 26262-1:2018}: Road vehicles - functional safety - part 1:
  Vocabulary, 2018.
\newblock URL \url{https://www.iso.org/standard/68383.html}.

\bibitem[Jermakian \& Zuby(2011)Jermakian and Zuby]{jermakian2011primary}
JS~Jermakian and DS~Zuby.
\newblock Primary pedestrian crash scenarios: factors relevant to the design of
  pedestrian detection systems.
\newblock \emph{Insurance Institute for Highway Safety, Arlington, VA}, 2011.

\bibitem[Ji et~al.(2019)Ji, Ren, and Law]{ji2019propagation}
Weiqi Ji, Zhuyin Ren, and Chung~K. Law.
\newblock Uncertainty propagation in deep neural network using active subspace.
\newblock \emph{CoRR}, abs/1903.03989, 2019.
\newblock URL \url{http://arxiv.org/abs/1903.03989}.

\bibitem[Kabir et~al.(2018)Kabir, Khosravi, Hosen, and
  Nahavandi]{kabir2018neural}
Hussain Mohammed~Dipu Kabir, Abbas Khosravi, Mohammad~Anwar Hosen, and Saeid
  Nahavandi.
\newblock Neural network-based uncertainty quantification: {A} survey of
  methodologies and applications.
\newblock \emph{{IEEE} Access}, 6:\penalty0 36218--36234, 2018.
\newblock \doi{10.1109/ACCESS.2018.2836917}.
\newblock URL \url{https://doi.org/10.1109/ACCESS.2018.2836917}.

\bibitem[Kamp et~al.(2018)Kamp, Adilova, Sicking, H{\"{u}}ger, Schlicht, Wirtz,
  and Wrobel]{kamp2018federated}
Michael Kamp, Linara Adilova, Joachim Sicking, Fabian H{\"{u}}ger, Peter
  Schlicht, Tim Wirtz, and Stefan Wrobel.
\newblock Efficient decentralized deep learning by dynamic model averaging.
\newblock In Michele Berlingerio, Francesco Bonchi, Thomas G{\"{a}}rtner, Neil
  Hurley, and Georgiana Ifrim (eds.), \emph{Machine Learning and Knowledge
  Discovery in Databases - European Conference, {ECML} {PKDD} 2018, Dublin,
  Ireland, September 10-14, 2018, Proceedings, Part {I}}, volume 11051 of
  \emph{Lecture Notes in Computer Science}, pp.\  393--409. Springer, 2018.
\newblock \doi{10.1007/978-3-030-10925-7\_24}.
\newblock URL \url{https://doi.org/10.1007/978-3-030-10925-7\_24}.

\bibitem[Kar et~al.(2019)Kar, Prakash, Liu, Cameracci, Yuan, Rusiniak, Acuna,
  Torralba, and Fidler]{kar2019meta}
Amlan Kar, Aayush Prakash, Ming{-}Yu Liu, Eric Cameracci, Justin Yuan, Matt
  Rusiniak, David Acuna, Antonio Torralba, and Sanja Fidler.
\newblock {Meta-Sim}: Learning to generate synthetic datasets.
\newblock In \emph{2019 {IEEE/CVF} International Conference on Computer Vision,
  {ICCV} 2019, Seoul, Korea (South), October 27 - November 2, 2019}, pp.\
  4550--4559. {IEEE}, 2019.
\newblock \doi{10.1109/ICCV.2019.00465}.
\newblock URL \url{https://doi.org/10.1109/ICCV.2019.00465}.

\bibitem[Kavitha et~al.(2010)Kavitha, Kavitha, and
  Suresh~Kumar]{kavitha2010requirementbased}
R.~Kavitha, V.R. Kavitha, and N.~Suresh~Kumar.
\newblock Requirement based test case prioritization.
\newblock In \emph{2010 INTERNATIONAL CONFERENCE ON COMMUNICATION CONTROL AND
  COMPUTING TECHNOLOGIES}, pp.\  826--829, 2010.
\newblock \doi{10.1109/ICCCCT.2010.5670728}.

\bibitem[Kelly \& Weaver(2004)Kelly and Weaver]{kelly2004goal}
Tim Kelly and Rob Weaver.
\newblock The goal structuring notation--a safety argument notation.
\newblock In \emph{Proceedings of the dependable systems and networks 2004
  workshop on assurance cases}, pp.\ ~6. Citeseer, 2004.

\bibitem[Kendall \& Gal(2017)Kendall and Gal]{kendall2017uncertainties}
Alex Kendall and Yarin Gal.
\newblock What uncertainties do we need in {B}ayesian deep learning for
  computer vision?
\newblock In Isabelle Guyon, Ulrike von Luxburg, Samy Bengio, Hanna~M. Wallach,
  Rob Fergus, S.~V.~N. Vishwanathan, and Roman Garnett (eds.), \emph{Advances
  in Neural Information Processing Systems 30: Annual Conference on Neural
  Information Processing Systems 2017, December 4-9, 2017, Long Beach, CA,
  {USA}}, pp.\  5574--5584, 2017.
\newblock URL
  \url{https://proceedings.neurips.cc/paper/2017/hash/2650d6089a6d640c5e85b2b88265dc2b-Abstract.html}.

\bibitem[Kim et~al.(2018)Kim, Wattenberg, Gilmer, Cai, Wexler, Vi{\'{e}}gas,
  and Sayres]{kim2018interpretability}
Been Kim, Martin Wattenberg, Justin Gilmer, Carrie~J. Cai, James Wexler,
  Fernanda~B. Vi{\'{e}}gas, and Rory Sayres.
\newblock Interpretability beyond feature attribution: Quantitative testing
  with concept activation vectors {(TCAV)}.
\newblock In Jennifer~G. Dy and Andreas Krause (eds.), \emph{Proceedings of the
  35th International Conference on Machine Learning, {ICML} 2018,
  Stockholmsm{\"{a}}ssan, Stockholm, Sweden, July 10-15, 2018}, volume~80 of
  \emph{Proceedings of Machine Learning Research}, pp.\  2673--2682. {PMLR},
  2018.
\newblock URL \url{http://proceedings.mlr.press/v80/kim18d.html}.

\bibitem[Klein et~al.(2009)Klein, Elphinstone, Heiser, Andronick, Cock, Derrin,
  Elkaduwe, Engelhardt, Kolanski, Norrish, Sewell, Tuch, and
  Winwood]{klein2009verifyoskernel}
Gerwin Klein, Kevin Elphinstone, Gernot Heiser, June Andronick, David Cock,
  Philip Derrin, Dhammika Elkaduwe, Kai Engelhardt, Rafal Kolanski, Michael
  Norrish, Thomas Sewell, Harvey Tuch, and Simon Winwood.
\newblock sel4: formal verification of an {OS} kernel.
\newblock In Jeanna~Neefe Matthews and Thomas~E. Anderson (eds.),
  \emph{Proceedings of the 22nd {ACM} Symposium on Operating Systems Principles
  2009, {SOSP} 2009, Big Sky, Montana, USA, October 11-14, 2009}, pp.\
  207--220. {ACM}, 2009.
\newblock \doi{10.1145/1629575.1629596}.
\newblock URL \url{https://doi.org/10.1145/1629575.1629596}.

\bibitem[Klischat \& Althoff(2019)Klischat and Althoff]{klischat2019scenarios}
Moritz Klischat and Matthias Althoff.
\newblock Generating critical test scenarios for automated vehicles with
  evolutionary algorithms.
\newblock In \emph{2019 {IEEE} Intelligent Vehicles Symposium, {IV} 2019,
  Paris, France, June 9-12, 2019}, pp.\  2352--2358. {IEEE}, 2019.
\newblock \doi{10.1109/IVS.2019.8814230}.
\newblock URL \url{https://doi.org/10.1109/IVS.2019.8814230}.

\bibitem[Kohl et~al.(2018)Kohl, Romera{-}Paredes, Meyer, Fauw, Ledsam,
  Maier{-}Hein, Eslami, Rezende, and Ronneberger]{kohl2018probabilistic}
Simon Kohl, Bernardino Romera{-}Paredes, Clemens Meyer, Jeffrey~De Fauw,
  Joseph~R. Ledsam, Klaus~H. Maier{-}Hein, S.~M.~Ali Eslami, Danilo~Jimenez
  Rezende, and Olaf Ronneberger.
\newblock A probabilistic u-net for segmentation of ambiguous images.
\newblock In Samy Bengio, Hanna~M. Wallach, Hugo Larochelle, Kristen Grauman,
  Nicol{\`{o}} Cesa{-}Bianchi, and Roman Garnett (eds.), \emph{Advances in
  Neural Information Processing Systems 31: Annual Conference on Neural
  Information Processing Systems 2018, NeurIPS 2018, December 3-8, 2018,
  Montr{\'{e}}al, Canada}, pp.\  6965--6975, 2018.
\newblock URL
  \url{https://proceedings.neurips.cc/paper/2018/hash/473447ac58e1cd7e96172575f48dca3b-Abstract.html}.

\bibitem[Koopman(2018)]{koopman2018heavy}
Philip Koopman.
\newblock The heavy tail safety ceiling.
\newblock In \emph{Automated and Connected Vehicle Systems Testing Symposium},
  volume 1145, 2018.

\bibitem[Koopman \& Fratrik(2019)Koopman and Fratrik]{koopman2019odd}
Philip Koopman and Frank Fratrik.
\newblock How many operational design domains, objects, and events?
\newblock In Hu{\'{a}}scar Espinoza, Se{\'{a}}n~{\'{O}} h{\'{E}}igeartaigh,
  Xiaowei Huang, Jos{\'{e}} Hern{\'{a}}ndez{-}Orallo, and Mauricio
  Castillo{-}Effen (eds.), \emph{Workshop on Artificial Intelligence Safety
  2019 co-located with the Thirty-Third {AAAI} Conference on Artificial
  Intelligence 2019 (AAAI-19), Honolulu, Hawaii, January 27, 2019}, volume 2301
  of \emph{{CEUR} Workshop Proceedings}. CEUR-WS.org, 2019.
\newblock URL \url{http://ceur-ws.org/Vol-2301/paper\_6.pdf}.

\bibitem[Koopman \& Osyk(2019)Koopman and Osyk]{koopman2019safetyac}
Philip Koopman and Beth Osyk.
\newblock Safety argument considerations for public road testing of autonomous
  vehicles.
\newblock \emph{SAE Technical Paper Series}, 2019.

\bibitem[Koopman et~al.(2019)Koopman, Ferrell, Fratrik, and
  Wagner]{koopman2019standard}
Philip Koopman, Uma Ferrell, Frank Fratrik, and Michael~D. Wagner.
\newblock A safety standard approach for fully autonomous vehicles.
\newblock In Alexander~B. Romanovsky, Elena Troubitsyna, Ilir Gashi, Erwin
  Schoitsch, and Friedemann Bitsch (eds.), \emph{Computer Safety, Reliability,
  and Security - {SAFECOMP} 2019 Workshops, ASSURE, DECSoS, SASSUR, STRIVE, and
  WAISE, Turku, Finland, September 10, 2019, Proceedings}, volume 11699 of
  \emph{Lecture Notes in Computer Science}, pp.\  326--332. Springer, 2019.
\newblock \doi{10.1007/978-3-030-26250-1\_26}.
\newblock URL \url{https://doi.org/10.1007/978-3-030-26250-1\_26}.

\bibitem[Kropp et~al.(1998)Kropp, Jr., and Siewiorek]{kropp1998robusttesting}
Nathan~P. Kropp, Philip J.~Koopman Jr., and Daniel~P. Siewiorek.
\newblock Automated robustness testing of off-the-shelf software components.
\newblock In \emph{Digest of Papers: FTCS-28, The Twenty-Eigth Annual
  International Symposium on Fault-Tolerant Computing, Munich, Germany, June
  23-25, 1998}, pp.\  230--239. {IEEE} Computer Society, 1998.
\newblock \doi{10.1109/FTCS.1998.689474}.
\newblock URL \url{https://doi.org/10.1109/FTCS.1998.689474}.

\bibitem[Kuleshov et~al.(2018)Kuleshov, Fenner, and Ermon]{Kuleshov2018}
Volodymyr Kuleshov, Nathan Fenner, and Stefano Ermon.
\newblock Accurate uncertainties for deep learning using calibrated regression.
\newblock In Jennifer Dy and Andreas Krause (eds.), \emph{Proceedings of the
  35th International Conference on Machine Learning}, volume~80 of
  \emph{Proceedings of Machine Learning Research}, pp.\  2796--2804. PMLR,
  10--15 Jul 2018.
\newblock URL \url{http://proceedings.mlr.press/v80/kuleshov18a.html}.

\bibitem[Kumar et~al.(2019)Kumar, Liang, and Ma]{kumar2019verified}
Ananya Kumar, Percy Liang, and Tengyu Ma.
\newblock Verified uncertainty calibration.
\newblock In Hanna~M. Wallach, Hugo Larochelle, Alina Beygelzimer, Florence
  d'Alch{\'{e}}{-}Buc, Emily~B. Fox, and Roman Garnett (eds.), \emph{Advances
  in Neural Information Processing Systems 32: Annual Conference on Neural
  Information Processing Systems 2019, NeurIPS 2019, December 8-14, 2019,
  Vancouver, BC, Canada}, pp.\  3787--3798, 2019.
\newblock URL
  \url{https://proceedings.neurips.cc/paper/2019/hash/f8c0c968632845cd133308b1a494967f-Abstract.html}.

\bibitem[Lakshminarayanan et~al.(2017)Lakshminarayanan, Pritzel, and
  Blundell]{lakshminarayanan2017simple}
Balaji Lakshminarayanan, Alexander Pritzel, and Charles Blundell.
\newblock Simple and scalable predictive uncertainty estimation using deep
  ensembles.
\newblock In Isabelle Guyon, Ulrike von Luxburg, Samy Bengio, Hanna~M. Wallach,
  Rob Fergus, S.~V.~N. Vishwanathan, and Roman Garnett (eds.), \emph{Advances
  in Neural Information Processing Systems 30: Annual Conference on Neural
  Information Processing Systems 2017, December 4-9, 2017, Long Beach, CA,
  {USA}}, pp.\  6402--6413, 2017.
\newblock URL
  \url{https://proceedings.neurips.cc/paper/2017/hash/9ef2ed4b7fd2c810847ffa5fa85bce38-Abstract.html}.

\bibitem[Lewis \& Gale(1994)Lewis and Gale]{lewis1994text}
David~D. Lewis and William~A. Gale.
\newblock A sequential algorithm for training text classifiers.
\newblock In W.~Bruce Croft and C.~J. van Rijsbergen (eds.), \emph{Proceedings
  of the 17th Annual International {ACM-SIGIR} Conference on Research and
  Development in Information Retrieval. Dublin, Ireland, 3-6 July 1994 (Special
  Issue of the {SIGIR} Forum)}, pp.\  3--12. ACM/Springer, 1994.
\newblock \doi{10.1007/978-1-4471-2099-5\_1}.
\newblock URL \url{https://doi.org/10.1007/978-1-4471-2099-5\_1}.

\bibitem[Liu et~al.(2019)Liu, Paisley, Kioumourtzoglou, and
  Coull]{liu2019accurate}
Jeremiah~Z. Liu, John~W. Paisley, Marianthi{-}Anna Kioumourtzoglou, and
  Brent~A. Coull.
\newblock Accurate uncertainty estimation and decomposition in ensemble
  learning.
\newblock In Hanna~M. Wallach, Hugo Larochelle, Alina Beygelzimer, Florence
  d'Alch{\'{e}}{-}Buc, Emily~B. Fox, and Roman Garnett (eds.), \emph{Advances
  in Neural Information Processing Systems 32: Annual Conference on Neural
  Information Processing Systems 2019, NeurIPS 2019, December 8-14, 2019,
  Vancouver, BC, Canada}, pp.\  8950--8961, 2019.
\newblock URL
  \url{https://proceedings.neurips.cc/paper/2019/hash/1cc8a8ea51cd0adddf5dab504a285915-Abstract.html}.

\bibitem[Lwakatare et~al.(2020)Lwakatare, Raj, Crnkovic, Bosch, and
  Olsson]{lwakatare2020large}
Lucy~Ellen Lwakatare, Aiswarya Raj, Ivica Crnkovic, Jan Bosch, and
  Helena~Holmstr{\"{o}}m Olsson.
\newblock Large-scale machine learning systems in real-world industrial
  settings: {A} review of challenges and solutions.
\newblock \emph{Inf. Softw. Technol.}, 127:\penalty0 106368, 2020.
\newblock \doi{10.1016/j.infsof.2020.106368}.
\newblock URL \url{https://doi.org/10.1016/j.infsof.2020.106368}.

\bibitem[Maddox et~al.(2019)Maddox, Izmailov, Garipov, Vetrov, and
  Wilson]{maddox2019simple}
Wesley~J Maddox, Pavel Izmailov, Timur Garipov, Dmitry~P Vetrov, and
  Andrew~Gordon Wilson.
\newblock A simple baseline for {B}ayesian uncertainty in deep learning.
\newblock In H.~Wallach, H.~Larochelle, A.~Beygelzimer, F.~d\textquotesingle
  Alch\'{e}-Buc, E.~Fox, and R.~Garnett (eds.), \emph{Advances in Neural
  Information Processing Systems}, volume~32. Curran Associates, Inc., 2019.
\newblock URL
  \url{https://proceedings.neurips.cc/paper/2019/file/118921efba23fc329e6560b27861f0c2-Paper.pdf}.

\bibitem[Madry et~al.(2018)Madry, Makelov, Schmidt, Tsipras, and
  Vladu]{madry2017towards}
Aleksander Madry, Aleksandar Makelov, Ludwig Schmidt, Dimitris Tsipras, and
  Adrian Vladu.
\newblock Towards deep learning models resistant to adversarial attacks.
\newblock In \emph{6th International Conference on Learning Representations,
  {ICLR} 2018, Vancouver, BC, Canada, April 30 - May 3, 2018, Conference Track
  Proceedings}. OpenReview.net, 2018.
\newblock URL \url{https://openreview.net/forum?id=rJzIBfZAb}.

\bibitem[Majzoobi et~al.(2008)Majzoobi, Koushanfar, and
  Potkonjak]{majzoobi2008hardwaresecurity}
Mehrdad Majzoobi, Farinaz Koushanfar, and Miodrag Potkonjak.
\newblock Testing techniques for hardware security.
\newblock In Douglas Young and Nur~A. Touba (eds.), \emph{2008 {IEEE}
  International Test Conference, {ITC} 2008, Santa Clara, California, USA,
  October 26-31, 2008}, pp.\  1--10. {IEEE} Computer Society, 2008.
\newblock \doi{10.1109/TEST.2008.4700636}.
\newblock URL \url{https://doi.org/10.1109/TEST.2008.4700636}.

\bibitem[Malinin \& Gales(2018)Malinin and Gales]{malinin2018predictive}
Andrey Malinin and Mark J.~F. Gales.
\newblock Predictive uncertainty estimation via prior networks.
\newblock In Samy Bengio, Hanna~M. Wallach, Hugo Larochelle, Kristen Grauman,
  Nicol{\`{o}} Cesa{-}Bianchi, and Roman Garnett (eds.), \emph{Advances in
  Neural Information Processing Systems 31: Annual Conference on Neural
  Information Processing Systems 2018, NeurIPS 2018, December 3-8, 2018,
  Montr{\'{e}}al, Canada}, pp.\  7047--7058, 2018.
\newblock URL
  \url{https://proceedings.neurips.cc/paper/2018/hash/3ea2db50e62ceefceaf70a9d9a56a6f4-Abstract.html}.

\bibitem[McMahan et~al.(2017)McMahan, Moore, Ramage, Hampson, and
  y~Arcas]{mcmahan2017federated}
Brendan McMahan, Eider Moore, Daniel Ramage, Seth Hampson, and
  Blaise~Ag{\"{u}}era y~Arcas.
\newblock Communication-efficient learning of deep networks from decentralized
  data.
\newblock In Aarti Singh and Xiaojin~(Jerry) Zhu (eds.), \emph{Proceedings of
  the 20th International Conference on Artificial Intelligence and Statistics,
  {AISTATS} 2017, 20-22 April 2017, Fort Lauderdale, FL, {USA}}, volume~54 of
  \emph{Proceedings of Machine Learning Research}, pp.\  1273--1282. {PMLR},
  2017.
\newblock URL \url{http://proceedings.mlr.press/v54/mcmahan17a.html}.

\bibitem[Meth et~al.(2015)Meth, M{\"{u}}ller, and
  Maedche]{meth2015designingrequirement}
Hendrik Meth, Benjamin M{\"{u}}ller, and Alexander Maedche.
\newblock Designing a requirement mining system.
\newblock \emph{J. Assoc. Inf. Syst.}, 16\penalty0 (9):\penalty0 2, 2015.
\newblock URL \url{http://aisel.aisnet.org/jais/vol16/iss9/2}.

\bibitem[Mitchell et~al.(2019)Mitchell, Wu, Zaldivar, Barnes, Vasserman,
  Hutchinson, Spitzer, Raji, and Gebru]{mitchell2019cards}
Margaret Mitchell, Simone Wu, Andrew Zaldivar, Parker Barnes, Lucy Vasserman,
  Ben Hutchinson, Elena Spitzer, Inioluwa~Deborah Raji, and Timnit Gebru.
\newblock Model cards for model reporting.
\newblock In danah boyd and Jamie~H. Morgenstern (eds.), \emph{Proceedings of
  the Conference on Fairness, Accountability, and Transparency, FAT* 2019,
  Atlanta, GA, USA, January 29-31, 2019}, pp.\  220--229. {ACM}, 2019.
\newblock \doi{10.1145/3287560.3287596}.
\newblock URL \url{https://doi.org/10.1145/3287560.3287596}.

\bibitem[Mock et~al.(2021)Mock, Scholz, Blank, H{\"{u}}ger, Rohatschek,
  Schwarz, and Stauner]{mock2021safetyargumentation}
Michael Mock, Stephan Scholz, Fr{\'{e}}d{\'{e}}rik Blank, Fabian H{\"{u}}ger,
  Andreas~J. Rohatschek, Loren Schwarz, and Thomas Stauner.
\newblock An integrated approach to a safety argumentation for ai-based
  perception functions in automated driving.
\newblock In Ibrahim Habli, Mark Sujan, Simos Gerasimou, Erwin Schoitsch, and
  Friedemann Bitsch (eds.), \emph{Computer Safety, Reliability, and Security.
  {SAFECOMP} 2021 Workshops - DECSoS, MAPSOD, DepDevOps, USDAI, and WAISE,
  York, UK, September 7, 2021, Proceedings}, volume 12853 of \emph{Lecture
  Notes in Computer Science}, pp.\  265--271. Springer, 2021.
\newblock \doi{10.1007/978-3-030-83906-2\_21}.
\newblock URL \url{https://doi.org/10.1007/978-3-030-83906-2\_21}.

\bibitem[Naeini et~al.(2015)Naeini, Cooper, and
  Hauskrecht]{naeini2015calibrated}
Mahdi~Pakdaman Naeini, Gregory~F. Cooper, and Milos Hauskrecht.
\newblock Obtaining well calibrated probabilities using bayesian binning.
\newblock In Blai Bonet and Sven Koenig (eds.), \emph{Proceedings of the
  Twenty-Ninth {AAAI} Conference on Artificial Intelligence, January 25-30,
  2015, Austin, Texas, {USA}}, pp.\  2901--2907. {AAAI} Press, 2015.
\newblock URL
  \url{http://www.aaai.org/ocs/index.php/AAAI/AAAI15/paper/view/9667}.

\bibitem[Nebut et~al.(2003)Nebut, Fleurey, Traon, and
  J{\'{e}}z{\'{e}}quel]{nebut2003productfamilies}
Cl{\'{e}}mentine Nebut, Franck Fleurey, Yves~Le Traon, and Jean{-}Marc
  J{\'{e}}z{\'{e}}quel.
\newblock A requirement-based approach to test product families.
\newblock In Frank van~der Linden (ed.), \emph{Software Product-Family
  Engineering, 5th International Workshop, {PFE} 2003, Siena, Italy, November
  4-6, 2003, Revised Papers}, volume 3014 of \emph{Lecture Notes in Computer
  Science}, pp.\  198--210. Springer, 2003.
\newblock \doi{10.1007/978-3-540-24667-1\_15}.
\newblock URL \url{https://doi.org/10.1007/978-3-540-24667-1\_15}.

\bibitem[NHTSA(2017)]{nhtsa2017adsafety}
NHTSA.
\newblock Automated driving systems 2.0: A vision for safety, 2017.
\newblock URL
  \url{https://www.nhtsa.gov/sites/nhtsa.gov/files/documents/13069a-ads2.0_090617_v9a_tag.pdf}.

\bibitem[Nix \& Weigend(1994)Nix and Weigend]{nix1994estimating}
D.A. Nix and A.S. Weigend.
\newblock Estimating the mean and variance of the target probability
  distribution.
\newblock In \emph{Proceedings of 1994 IEEE International Conference on Neural
  Networks (ICNN'94)}, volume~1, pp.\  55--60 vol.1, 1994.
\newblock \doi{10.1109/ICNN.1994.374138}.

\bibitem[Odena et~al.(2019)Odena, Olsson, Andersen, and
  Goodfellow]{odena2019tensorfuzz}
Augustus Odena, Catherine Olsson, David~G. Andersen, and Ian~J. Goodfellow.
\newblock Tensorfuzz: Debugging neural networks with coverage-guided fuzzing.
\newblock In Kamalika Chaudhuri and Ruslan Salakhutdinov (eds.),
  \emph{Proceedings of the 36th International Conference on Machine Learning,
  {ICML} 2019, 9-15 June 2019, Long Beach, California, {USA}}, volume~97 of
  \emph{Proceedings of Machine Learning Research}, pp.\  4901--4911. {PMLR},
  2019.
\newblock URL \url{http://proceedings.mlr.press/v97/odena19a.html}.

\bibitem[Pandey et~al.(2010)Pandey, Suman, and
  Ramani]{pandey2010requirementeng}
Dhirendra Pandey, U.~Suman, and A.K. Ramani.
\newblock An effective requirement engineering process model for software
  development and requirements management.
\newblock In \emph{2010 International Conference on Advances in Recent
  Technologies in Communication and Computing}, pp.\  287--291, 2010.
\newblock \doi{10.1109/ARTCom.2010.24}.

\bibitem[Papananias et~al.(2019)Papananias, McLeay, Mahfouf, and
  Kadirkamanathan]{papananias2019manufacturing}
Moschos Papananias, Thomas~E. McLeay, Mahdi Mahfouf, and Visakan
  Kadirkamanathan.
\newblock A bayesian framework to estimate part quality and associated
  uncertainties in multistage manufacturing.
\newblock \emph{Comput. Ind.}, 105:\penalty0 35--47, 2019.
\newblock \doi{10.1016/j.compind.2018.10.008}.
\newblock URL \url{https://doi.org/10.1016/j.compind.2018.10.008}.

\bibitem[Pei et~al.(2017)Pei, Cao, Yang, and Jana]{pei2017deepxplore}
Kexin Pei, Yinzhi Cao, Junfeng Yang, and Suman Jana.
\newblock Deepxplore: Automated whitebox testing of deep learning systems.
\newblock In \emph{Proceedings of the 26th Symposium on Operating Systems
  Principles, Shanghai, China, October 28-31, 2017}, pp.\  1--18. {ACM}, 2017.
\newblock \doi{10.1145/3132747.3132785}.
\newblock URL \url{https://doi.org/10.1145/3132747.3132785}.

\bibitem[Perneger(2005)]{perneger2005swiss}
Thomas Perneger.
\newblock The {Swiss} cheese model of safety incidents: Are there holes in the
  metaphor?
\newblock \emph{BMC health services research}, 5:\penalty0 71, 02 2005.
\newblock \doi{10.1186/1472-6963-5-71}.

\bibitem[Pintz et~al.(2022)Pintz, Sicking, Poretschkin, and
  Akila]{pintz2022survey-unc-toolkits}
Maximilian Pintz, Joachim Sicking, Maximilian Poretschkin, and Maram Akila.
\newblock A survey on uncertainty toolkits for deep learning.
\newblock \emph{ICLR 2022 Workshop on ML Evaluation Standards}, 2022.

\bibitem[Postels et~al.(2019)Postels, Ferroni, Coskun, Navab, and
  Tombari]{postels2019sampling}
Janis Postels, Francesco Ferroni, Huseyin Coskun, Nassir Navab, and Federico
  Tombari.
\newblock Sampling-free epistemic uncertainty estimation using approximated
  variance propagation.
\newblock In \emph{2019 {IEEE/CVF} International Conference on Computer Vision,
  {ICCV} 2019, Seoul, Korea (South), October 27 - November 2, 2019}, pp.\
  2931--2940. {IEEE}, 2019.
\newblock \doi{10.1109/ICCV.2019.00302}.
\newblock URL \url{https://doi.org/10.1109/ICCV.2019.00302}.

\bibitem[Reinkemeier et~al.(2011)Reinkemeier, Stierand, Rehkop, and
  Henkler]{reinkemeier2011requirement}
Philipp Reinkemeier, Ingo Stierand, Philip Rehkop, and Stefan Henkler.
\newblock A pattern-based requirement specification language: Mapping
  automotive specific timing requirements.
\newblock In Ralf~H. Reussner, Alexander Pretschner, and Stefan J{\"{a}}hnichen
  (eds.), \emph{Software Engineering 2011 - Workshopband (inkl.
  Doktorandensymposium), Fachtagung des GI-Fachbereichs Softwaretechnik,
  21.-25.02.2011, Karlsruhe}, volume {P-184} of \emph{{LNI}}, pp.\  99--108.
  {GI}, 2011.
\newblock URL \url{https://dl.gi.de/20.500.12116/19877}.

\bibitem[Riccio et~al.(2020)Riccio, Jahangirova, Stocco, Humbatova, Weiss, and
  Tonella]{riccio2020testing}
Vincenzo Riccio, Gunel Jahangirova, Andrea Stocco, Nargiz Humbatova, Michael
  Weiss, and Paolo Tonella.
\newblock Testing machine learning based systems: a systematic mapping.
\newblock \emph{Empir. Softw. Eng.}, 25\penalty0 (6):\penalty0 5193--5254,
  2020.
\newblock \doi{10.1007/s10664-020-09881-0}.
\newblock URL \url{https://doi.org/10.1007/s10664-020-09881-0}.

\bibitem[Rockafellar \& Uryasev(2002)Rockafellar and
  Uryasev]{rockafellar2002conditional}
R.Tyrrell Rockafellar and Stanislav Uryasev.
\newblock Conditional value-at-risk for general loss distributions.
\newblock \emph{Journal of Banking \& Finance}, 26\penalty0 (7):\penalty0
  1443--1471, 2002.
\newblock ISSN 0378-4266.
\newblock \doi{https://doi.org/10.1016/S0378-4266(02)00271-6}.
\newblock URL
  \url{https://www.sciencedirect.com/science/article/pii/S0378426602002716}.

\bibitem[Roy et~al.(2010)Roy, Kim, and Trivedi]{roy2010cyber}
Arpan Roy, Dong~Seong Kim, and Kishor~S. Trivedi.
\newblock Cyber security analysis using attack countermeasure trees.
\newblock In Frederick~T. Sheldon, Stacy~J. Prowell, Robert~K. Abercrombie, and
  Axel~W. Krings (eds.), \emph{Proceedings of the 6th Cyber Security and
  Information Intelligence Research Workshop, {CSIIRW} 2010, Oak Ridge, TN,
  USA, April 21-23, 2010}, pp.\ ~28. {ACM}, 2010.
\newblock \doi{10.1145/1852666.1852698}.
\newblock URL \url{https://doi.org/10.1145/1852666.1852698}.

\bibitem[Salman et~al.(2019)Salman, Yang, Zhang, Hsieh, and
  Zhang]{salman2019convexrelaxation}
Hadi Salman, Greg Yang, Huan Zhang, Cho{-}Jui Hsieh, and Pengchuan Zhang.
\newblock A convex relaxation barrier to tight robustness verification of
  neural networks.
\newblock In Hanna~M. Wallach, Hugo Larochelle, Alina Beygelzimer, Florence
  d'Alch{\'{e}}{-}Buc, Emily~B. Fox, and Roman Garnett (eds.), \emph{Advances
  in Neural Information Processing Systems 32: Annual Conference on Neural
  Information Processing Systems 2019, NeurIPS 2019, December 8-14, 2019,
  Vancouver, BC, Canada}, pp.\  9832--9842, 2019.
\newblock URL
  \url{https://proceedings.neurips.cc/paper/2019/hash/246a3c5544feb054f3ea718f61adfa16-Abstract.html}.

\bibitem[Scanlon et~al.(2021)Scanlon, Kusano, Daniel, Alderson, Ogle, and
  Victor]{scanlon2021waymo}
John~M Scanlon, Kristofer~D Kusano, Tom Daniel, Christopher Alderson, Alexander
  Ogle, and Trent Victor.
\newblock Waymo simulated driving behavior in reconstructed fatal crashes
  within an autonomous vehicle operating domain, 2021.

\bibitem[Sculley et~al.(2015)Sculley, Holt, Golovin, Davydov, Phillips, Ebner,
  Chaudhary, Young, Crespo, and Dennison]{sculley2015hidden}
D.~Sculley, Gary Holt, Daniel Golovin, Eugene Davydov, Todd Phillips, Dietmar
  Ebner, Vinay Chaudhary, Michael Young, Jean{-}Fran{\c{c}}ois Crespo, and Dan
  Dennison.
\newblock Hidden technical debt in machine learning systems.
\newblock In Corinna Cortes, Neil~D. Lawrence, Daniel~D. Lee, Masashi Sugiyama,
  and Roman Garnett (eds.), \emph{Advances in Neural Information Processing
  Systems 28: Annual Conference on Neural Information Processing Systems 2015,
  December 7-12, 2015, Montreal, Quebec, Canada}, pp.\  2503--2511, 2015.
\newblock URL
  \url{https://proceedings.neurips.cc/paper/2015/hash/86df7dcfd896fcaf2674f757a2463eba-Abstract.html}.

\bibitem[Selvaraju et~al.(2017)Selvaraju, Cogswell, Das, Vedantam, Parikh, and
  Batra]{selvaraju2017grad}
Ramprasaath~R. Selvaraju, Michael Cogswell, Abhishek Das, Ramakrishna Vedantam,
  Devi Parikh, and Dhruv Batra.
\newblock {Grad-CAM}: Visual explanations from deep networks via gradient-based
  localization.
\newblock In \emph{{IEEE} International Conference on Computer Vision, {ICCV}
  2017, Venice, Italy, October 22-29, 2017}, pp.\  618--626. {IEEE} Computer
  Society, 2017.
\newblock \doi{10.1109/ICCV.2017.74}.
\newblock URL \url{https://doi.org/10.1109/ICCV.2017.74}.

\bibitem[Shafaei et~al.(2018)Shafaei, Kugele, Osman, and
  Knoll]{shafaei2018safety}
Sina Shafaei, Stefan Kugele, Mohd~Hafeez Osman, and Alois~C. Knoll.
\newblock Uncertainty in machine learning: {A} safety perspective on autonomous
  driving.
\newblock In Barbara Gallina, Amund Skavhaug, Erwin Schoitsch, and Friedemann
  Bitsch (eds.), \emph{Computer Safety, Reliability, and Security - {SAFECOMP}
  2018 Workshops, ASSURE, DECSoS, SASSUR, STRIVE, and WAISE,
  V{\"{a}}ster{\aa}s, Sweden, September 18, 2018, Proceedings}, volume 11094 of
  \emph{Lecture Notes in Computer Science}, pp.\  458--464. Springer, 2018.
\newblock \doi{10.1007/978-3-319-99229-7\_39}.
\newblock URL \url{https://doi.org/10.1007/978-3-319-99229-7\_39}.

\bibitem[Sicking et~al.(2020)Sicking, Akila, Wirtz, Houben, and
  Fischer]{sicking2020characteristics}
Joachim Sicking, Maram Akila, Tim Wirtz, Sebastian Houben, and Asja Fischer.
\newblock Characteristics of {M}onte {C}arlo dropout in wide neural networks.
\newblock \emph{ICML 2020 Workshop on Uncertainty and Robustness in Deep
  Learning}, 2020.

\bibitem[Sicking et~al.(2021{\natexlab{a}})Sicking, Akila, Pintz, Wirtz,
  Fischer, and Wrobel]{sicking2021sml}
Joachim Sicking, Maram Akila, Maximilian Pintz, Tim Wirtz, Asja Fischer, and
  Stefan Wrobel.
\newblock A novel regression loss for non-parametric uncertainty optimization.
\newblock \emph{Symposium on Advances in Approximate Bayesian Inference
  (AABI)}, 3, 2021{\natexlab{a}}.
\newblock URL \url{https://openreview.net/forum?id=NHlpgYy-Sfq}.

\bibitem[Sicking et~al.(2021{\natexlab{b}})Sicking, Akila, Pintz, Wirtz,
  Wrobel, and Fischer]{sicking2021wdrop}
Joachim Sicking, Maram Akila, Maximilian Pintz, Tim Wirtz, Stefan Wrobel, and
  Asja Fischer.
\newblock Wasserstein dropout.
\newblock \emph{https://arxiv.org/abs/2012.12687}, 2021{\natexlab{b}}.

\bibitem[Snoek et~al.(2019)Snoek, Ovadia, Fertig, Lakshminarayanan, Nowozin,
  Sculley, Dillon, Ren, and Nado]{snoek2019can}
Jasper Snoek, Yaniv Ovadia, Emily Fertig, Balaji Lakshminarayanan, Sebastian
  Nowozin, D.~Sculley, Joshua~V. Dillon, Jie Ren, and Zachary Nado.
\newblock Can you trust your model's uncertainty? evaluating predictive
  uncertainty under dataset shift.
\newblock In Hanna~M. Wallach, Hugo Larochelle, Alina Beygelzimer, Florence
  d'Alch{\'{e}}{-}Buc, Emily~B. Fox, and Roman Garnett (eds.), \emph{Advances
  in Neural Information Processing Systems 32: Annual Conference on Neural
  Information Processing Systems 2019, NeurIPS 2019, December 8-14, 2019,
  Vancouver, BC, Canada}, pp.\  13969--13980, 2019.
\newblock URL
  \url{https://proceedings.neurips.cc/paper/2019/hash/8558cb408c1d76621371888657d2eb1d-Abstract.html}.

\bibitem[S{\"{u}}nderhauf et~al.(2018)S{\"{u}}nderhauf, Brock, Scheirer,
  Hadsell, Fox, Leitner, Upcroft, Abbeel, Burgard, Milford, and
  Corke]{sunderhauf2018dlforrobotics}
Niko S{\"{u}}nderhauf, Oliver Brock, Walter~J. Scheirer, Raia Hadsell, Dieter
  Fox, J{\"{u}}rgen Leitner, Ben Upcroft, Pieter Abbeel, Wolfram Burgard,
  Michael Milford, and Peter Corke.
\newblock The limits and potentials of deep learning for robotics.
\newblock \emph{Int. J. Robotics Res.}, 37\penalty0 (4-5):\penalty0 405--420,
  2018.
\newblock \doi{10.1177/0278364918770733}.
\newblock URL \url{https://doi.org/10.1177/0278364918770733}.

\bibitem[Tahat et~al.(2001)Tahat, Bader, Vaysburg, and
  Korel]{tahat2001reqbasedtestgeneration}
Luay~Ho Tahat, Atef Bader, Boris Vaysburg, and Bogdan Korel.
\newblock Requirement-based automated black-box test generation.
\newblock In \emph{25th International Computer Software and Applications
  Conference {(COMPSAC} 2001), Invigorating Software Development, 8-12 October
  2001, Chicago, IL, {USA}}, pp.\  489--495. {IEEE} Computer Society, 2001.
\newblock \doi{10.1109/CMPSAC.2001.960658}.
\newblock URL \url{https://doi.org/10.1109/CMPSAC.2001.960658}.

\bibitem[Thrun et~al.(2005)Thrun, Burgard, and Fox]{thrun2005robotics}
Sebastian Thrun, Wolfram Burgard, and Dieter Fox.
\newblock \emph{Probabilistic robotics}.
\newblock Intelligent robotics and autonomous agents. {MIT} Press, 2005.
\newblock ISBN 978-0-262-20162-9.

\bibitem[Truong{-}Le et~al.(2018)Truong{-}Le, Diehl, Brunner, and
  Knoll]{le2018odsafety}
Michael Truong{-}Le, Frederik Diehl, Thomas Brunner, and Alois~C. Knoll.
\newblock Uncertainty estimation for deep neural object detectors in
  safety-critical applications.
\newblock In Wei{-}Bin Zhang, Alexandre~M. Bayen, Javier J.~S{\'{a}}nchez
  Medina, and Matthew~J. Barth (eds.), \emph{21st International Conference on
  Intelligent Transportation Systems, {ITSC} 2018, Maui, HI, USA, November 4-7,
  2018}, pp.\  3873--3878. {IEEE}, 2018.
\newblock \doi{10.1109/ITSC.2018.8569637}.
\newblock URL \url{https://doi.org/10.1109/ITSC.2018.8569637}.

\bibitem[{Underwriters Laboratories}(2020)]{ul2020autonomous}
{Underwriters Laboratories}.
\newblock Standard for evaluation of autonomous products, 2020.
\newblock URL
  \url{https://www.shopulstandards.com/ProductDetail.aspx?productid=UL4600}.

\bibitem[Vapnik(2000)]{vapnik2000statistical}
Vladimir Vapnik.
\newblock \emph{The Nature of Statistical Learning Theory}.
\newblock Statistics for Engineering and Information Science. Springer, 2000.
\newblock ISBN 978-1-4419-3160-3.
\newblock \doi{10.1007/978-1-4757-3264-1}.
\newblock URL \url{https://doi.org/10.1007/978-1-4757-3264-1}.

\bibitem[Wahlster \& Winterhalter(2020)Wahlster and
  Winterhalter]{din2020roadmap}
Wolfgang Wahlster and Christoph Winterhalter.
\newblock German standardization roadmap on artificial intelligence, 2020.
\newblock URL
  \url{https://www.din.de/resource/blob/772610/e96c34dd6b12900ea75b460538805349/normungsroadmap-en-data.pdf}.

\bibitem[Welch \& Bishop(1995)Welch and Bishop]{welch1995introduction}
Greg Welch and Gary Bishop.
\newblock An introduction to the {K}alman filter.
\newblock Technical Report 95-041, University of North Carolina at Chapel Hill,
  Chapel Hill, NC, USA, 1995.
\newblock URL \url{http://www.cs.unc.edu/~welch/kalman/kalmanIntro.html}.

\bibitem[Welling \& Teh(2011)Welling and Teh]{welling2011bayesian}
Max Welling and Yee~Whye Teh.
\newblock {B}ayesian learning via stochastic gradient {L}angevin dynamics.
\newblock In Lise Getoor and Tobias Scheffer (eds.), \emph{Proceedings of the
  28th International Conference on Machine Learning, {ICML} 2011, Bellevue,
  Washington, USA, June 28 - July 2, 2011}, pp.\  681--688. Omnipress, 2011.
\newblock URL \url{https://icml.cc/2011/papers/398\_icmlpaper.pdf}.

\bibitem[Willers et~al.(2020)Willers, Sudholt, Raafatnia, and
  Abrecht]{willers2020safety}
Oliver Willers, Sebastian Sudholt, Shervin Raafatnia, and Stephanie Abrecht.
\newblock Safety concerns and mitigation approaches regarding the use of deep
  learning in safety-critical perception tasks.
\newblock In Ant{\'{o}}nio Casimiro, Frank Ortmeier, Erwin Schoitsch,
  Friedemann Bitsch, and Pedro~M. Ferreira (eds.), \emph{Computer Safety,
  Reliability, and Security. {SAFECOMP} 2020 Workshops - DECSoS 2020, DepDevOps
  2020, {USDAI} 2020, and {WAISE} 2020, Lisbon, Portugal, September 15, 2020,
  Proceedings}, volume 12235 of \emph{Lecture Notes in Computer Science}, pp.\
  336--350. Springer, 2020.
\newblock \doi{10.1007/978-3-030-55583-2\_25}.
\newblock URL \url{https://doi.org/10.1007/978-3-030-55583-2\_25}.

\bibitem[Wright et~al.(2000)Wright, Ramage, Cornford, and
  Nabney]{wright2000input}
W.~A. Wright, Guillaume Ramage, Dan Cornford, and Ian Nabney.
\newblock Neural network modelling with input uncertainty: Theory and
  application.
\newblock \emph{J. {VLSI} Signal Process.}, 26\penalty0 (1-2):\penalty0
  169--188, 2000.
\newblock \doi{10.1023/A:1008111920791}.
\newblock URL \url{https://doi.org/10.1023/A:1008111920791}.

\bibitem[Wu et~al.(2017)Wu, Iandola, Jin, and Keutzer]{wu2017squeezedet}
Bichen Wu, Forrest Iandola, Peter~H Jin, and Kurt Keutzer.
\newblock Squeeze{D}et: Unified, small, low power fully convolutional neural
  networks for real-time object detection for autonomous driving.
\newblock In \emph{Proceedings of the IEEE Conference on Computer Vision and
  Pattern Recognition Workshops}, pp.\  129--137, 2017.

\bibitem[Yao et~al.(2019)Yao, Xu, Wang, Crandall, and Atkins]{yao2019accidents}
Yu~Yao, Mingze Xu, Yuchen Wang, David~J. Crandall, and Ella~M. Atkins.
\newblock Unsupervised traffic accident detection in first-person videos.
\newblock In \emph{2019 {IEEE/RSJ} International Conference on Intelligent
  Robots and Systems, {IROS} 2019, Macau, SAR, China, November 3-8, 2019}, pp.\
   273--280. {IEEE}, 2019.
\newblock \doi{10.1109/IROS40897.2019.8967556}.
\newblock URL \url{https://doi.org/10.1109/IROS40897.2019.8967556}.

\bibitem[Zhang et~al.(2020)Zhang, Harman, Ma, and Liu]{zhang2020machine}
Jie~M. Zhang, Mark Harman, Lei Ma, and Yang Liu.
\newblock Machine learning testing: Survey, landscapes and horizons.
\newblock \emph{IEEE Transactions on Software Engineering}, pp.\  1--1, 2020.
\newblock \doi{10.1109/TSE.2019.2962027}.

\bibitem[Zhang \& Zhuang(2019)Zhang and Zhuang]{zhang2019attacker}
Jing Zhang and Jun Zhuang.
\newblock Modeling a multi-target attacker-defender game with multiple attack
  types.
\newblock \emph{Reliab. Eng. Syst. Saf.}, 185:\penalty0 465--475, 2019.
\newblock \doi{10.1016/j.ress.2019.01.015}.
\newblock URL \url{https://doi.org/10.1016/j.ress.2019.01.015}.

\end{thebibliography}
\bibliographystyle{iclr2021_conference}

\clearpage


\end{document}